\documentclass[a4paper,fleqn]{cas-dc}

\usepackage{hyperref}
\usepackage{amsmath,amssymb,amsfonts}
\usepackage[square, sort, numbers]{natbib}
\usepackage[ruled,vlined]{algorithm2e}
\SetKwInput{KwInput}{Input}
\SetKwInput{KwOutput}{Output}
\usepackage{caption}
\usepackage[labelformat=simple]{subcaption}

\usepackage{float}
\usepackage{multirow}
\usepackage{graphicx}
\usepackage{textcomp}
\usepackage{xcolor}
\usepackage{tabularx,booktabs}
\newcolumntype{Y}{>{\centering\arraybackslash}X}

\ifodd 1
\else
\fi

\begin{document}
\let\WriteBookmarks\relax
\def\floatpagepagefraction{1}
\def\textpagefraction{.001}

% \shorttitle{}

\shortauthors{This paper has been accepted by Robotics and Autonomous Systems with \href{https://doi.org/10.1016/j.robot.2024.104752}{DOI: 10.1016/j.robot.2024.104752}.}

\title[mode = title]{Distributed multi-robot potential-field-based exploration \\with submap-based mapping and noise-augmented strategy}

\affiliation[1]{organization={ Engineering Product Development, Singapore University of Technology and Design},
    country={Singapore 487372}}
    
\affiliation[2]{organization={ Inria Grenoble, Grenoble Alpes University},
    country={France 38330}}

\affiliation[3]{organization={ School of Electrical and Electronic Engineering, Nanyang Technological University},
    country={Singapore 639798}}

\author[1]{Khattiya Pongsirijinda}
\cormark[1]
\ead{khattiya_pongsirijinda@mymail.sutd.edu.sg}

\author[1]{Zhiqiang Cao}
\author[2]{Kaushik Bhowmik}
\author[1]{Muhammad Shalihan}
\author[1]{\\Billy Pik Lik Lau}
\author[3]{Ran Liu}
\author[3]{Chau Yuen}
\author[1]{U-Xuan Tan}

\cortext[cor1]{~Corresponding author.}

\begin{abstract}
Multi-robot collaboration has become a needed component in unknown environment exploration due to its ability to accomplish various challenging situations. Potential-field-based methods are widely used for autonomous exploration because of their high efficiency and low travel cost. However, exploration speed and collaboration ability are still challenging topics. Therefore, we propose a \underline{D}istributed \underline{M}ulti-Robot \underline{P}otential-\underline{F}ield-Based Exploration (DMPF-Explore). In particular, we first present a \underline{D}istributed \underline{S}ubmap-Based \underline{M}ulti-Robot \underline{C}ollaborative Mapping Method (DSMC-Map), which can efficiently estimate the robot trajectories and construct the global map by merging the local maps from each robot. Second, we introduce a Potential-Field-Based Exploration Strategy Augmented with \underline{M}odified \underline{W}ave-\underline{F}ront Distance and \underline{C}olored \underline{N}oises (MWF-CN), in which the accessible frontier neighborhood is extended, and the colored noise provokes the enhancement of exploration performance. The proposed exploration method is deployed for simulation and real-world scenarios. The results show that our approach outperforms the existing ones regarding exploration speed and collaboration ability.
\end{abstract}

% Keywords

\maketitle

\section{Introduction}

Multi-robot exploration has gained significant attention in recent decades due to its potential applications in various challenging domains, such as high-tech industries, search and rescue tasks, environmental monitoring, and subterranean exploration \cite{SurveySub}. It can reduce the risk of human exposure to exploration and navigation tasks in dangerous and inaccessible areas. Due to the functionality of multi-robot systems, there are recent studies in various aspects, such as multi-robot collaborative localization \cite{SLAM1, SLAM2}, navigation \cite{C-Nav, ShortNav}, path and motion planning \cite{PathPlanning1, PathPlanning2, PathPlanning3, MotionPlanning}, etc.

This paper will mainly focus on the exploration method, in which mapping and exploration strategy play principal roles. Both of them are fundamental functions in multi-robot exploration tasks, which will directly affect the performance of the overall multi-robot exploration system.
So far, most of the current map merging methods implement mapping and merging independently, in which each robot separately builds a map within the local coordinate system and merges the maps from the rest robots into a common reference frame \cite{SMMR, MapMerging}.
Since the two parts are completely independent, the quality and consistency of the merged map will not be guaranteed. This constraint can consequently impact exploration speed and inter-robot coordination.

Moving on to look at the multi-robot exploration strategy, we notice that exploration strategies with different techniques were proposed, for example, cost-based \cite{cost}, sensor-based \cite{Sensor-based}, sampling-based \cite{RRT, MultiRRT, Sampling-based, RRT3}, potential-field-based \cite{SMMR, APF}, reinforcement-learning-based \cite{DRL1, DRL2}, and hierarchical-based \cite{hier1, hier2}. It is rather challenging for the existing strategies to make the robots collaboratively and efficiently explore real-world unknown environments. Most of them are still based on greedy or heuristic manners. Some exploration methods \cite{3D1, 3D2, 3D3, 3D4} are devoted to only unmanned aerial vehicles (UAVs) and 3D environment representation. In addition, some methods do not focus on detecting the environment's corner details, which is necessary for rescue scenarios because robots can be trapped at small corners or narrow holes. Furthermore, some techniques do not have any adjustable parameters. These constraints make their exploration tend to be repetitious. So, if they cannot succeed in exploring the given environment in early iterations, they tend not to be able to accomplish it even after several attempts.

Recently, there were newly developed multi-robot exploration methods that improved robot performance in exploring unknown environments. One of the recent approaches applied the novel temporal-memory-based strategy \cite{TM-RRT} together with the conventional RRT (Rapidly-exploring Random Tree) exploration \cite{RRT, MultiRRT}. Another freshly proposed method \cite{MUI-TARE} concentrated on the situation when the robots' initial positions are unknown. However, these mentioned methods are still based on the centralized architecture. So, they have yet to satisfy another critical aspect of multi-robot exploration regarding the independent coordination structure between robots. Since a central coordinator will control and make the decisions for all the robots, communication and computation issues can arise. Recent years have gained research interests in distributed exploration manner, as it allows the exploration to continue functioning in the scenarios of lost communications and be more appropriate in environments where robots can be stuck more easily. Because if a robot malfunctions, other robots can still continue the exploration independently in the case of distributed solution. Some distributed strategies \cite{SMMR, MR-Topo} have been introduced for multi-robot exploration. However, there are still rooms for improvement due to their mapping ability and exploration strategy, especially in unknown environments of size more than 200$\text{m}^2$, which are considered large-scale, and also the complex settings, which are not simply polygonal structures, contain small details and involve several pathways.

\begin{figure}
    \centering
        \includegraphics[width=\linewidth]{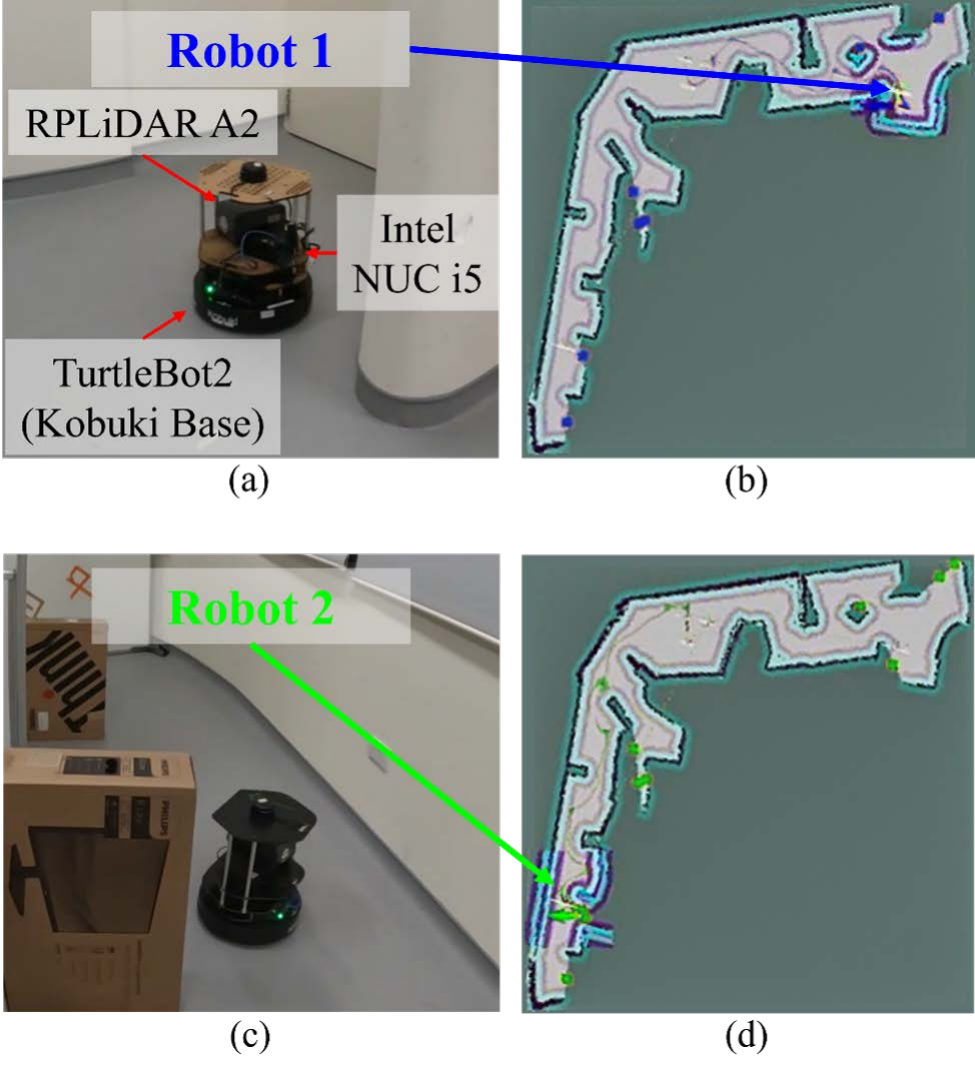}
        \vspace*{-6mm}
        \caption{Illustration of the DMPF-Explore with distributed mapping and exploration running on the individual robot. The map in Fig. 1(b) is built by Robot 1 in Fig. 1(a), and the map in Fig. 1(d) is built by Robot 2 in Fig. 1(c).}
        \vspace*{-5mm}
        \label{fig:turtlebot}
    \centering
\end{figure}

Therefore, in this paper, we propose a \textbf{D}istributed \textbf{M}ulti-Robot \textbf{P}otential-\textbf{F}ield-Based Exploration (DMPF-Explore) consisting of two main components: mapping and exploration. Firstly, we present a \textbf{D}istributed \textbf{S}ubmap-Based \textbf{M}ulti-Robot \textbf{C}ollaborative Mapping Method (DSMC-Map) to construct the occupancy map of the environment by combing the individual robot local maps, as shown in Fig. \ref{fig:turtlebot}. Our method is based on the submaps, which are the accumulation of scans over a time period. Loop closures, which are the processes of identifying previously visited locations, are also tackled in this part. In particular, the DSMC-Map comprises three modules: submap construction, loop closure detection (including intra-robot loop closures and inter-robot loop closures), and distributed pose graph optimization.
Secondly, another key element that we introduce is a potential-field-based exploration strategy augmented with \textbf{M}odified \textbf{W}ave-\textbf{F}ront distance and \textbf{C}olored \textbf{N}oises (MWF-CN). In particular, the attractive potential of frontiers with the modified wave-front (MWF) distance is for leading robots to appropriate goals. At the same time, the repulsive potential of robots fluctuated by the colored noise is also assigned during the exploration to diversify the potential values. In summary, the main contributions of our proposed method are as follows:

\begin{itemize}
    \item To deal with the reliability and consistency of the mapping, we propose the DSMC-Map combining LiDAR loop closures and odometry measurements in a novel manner. In particular, each robot performs distributed pose graph optimization to guarantee the consistency and accuracy of the pose estimation and build a global map including all areas explored by robots.
    \item To overcome challenging exploration scenarios, we propose the MWF-CN consisting of the attractive potential of frontiers and the repulsive potential of robots. In particular, the MWF distance helps to boost the attractive potential in a new way by extending the frontier neighborhood and hindering the local optima problems. The colored noise also novelly exerts diverse repulsive potential, leading to better exploration performance.
    \item We implement the DMPF-Explore and test it in simulated and real-world unknown environments. Our approach is compared with a benchmark to show its superiority in exploration efficiency and collaboration ability.
\end{itemize}

The remaining sections of this paper proceed as follows: Firstly, in Section \ref{section:mapping}, the DSMC-Map is described in three parts, namely submap construction, loop closure detection, and distributed pose graph optimization. Then, the attractive and repulsive potentials from the exploration strategy, MWF-CN, will be explained in Section \ref{section:exploration}. In Section \ref{section:simulation}, we will provide the details about the evaluation metrics and explain how the simulation is conducted. After that, the simulation results are also analyzed. Next, in Section \ref{section:real-world}, we demonstrate how the real-world deployment is designed and discuss the experimental results. And lastly, in Section \ref{section:conclusion}, we conclude the main findings and identify areas for future research.

\section{DSMC-Map mapping strategy}
\label{section:mapping}

\begin{figure}
  \begin{center}
  \includegraphics[width=\linewidth]{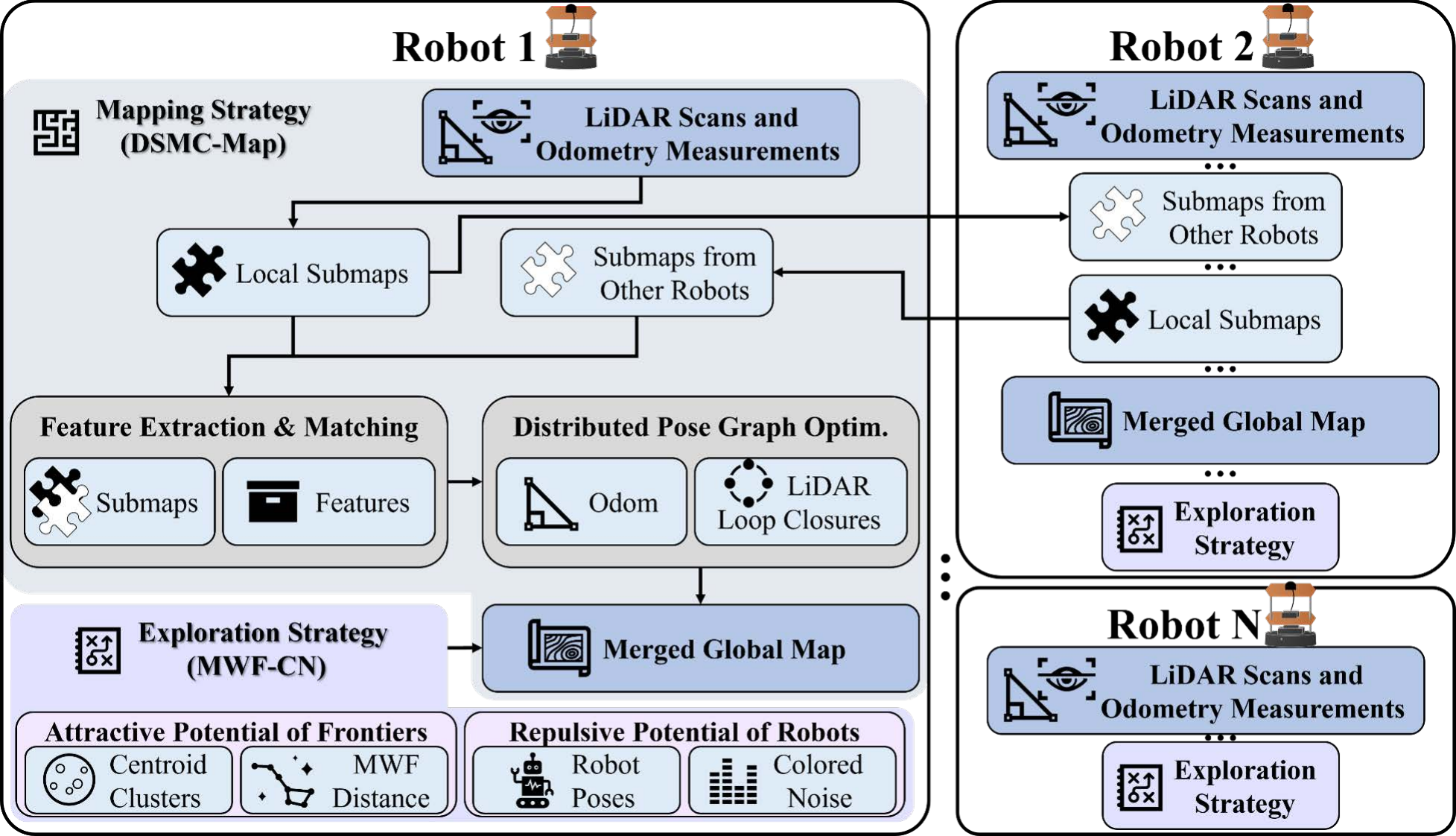}
  \vspace*{-5mm}
  \caption{Overview of the DMPF-Explore framework}
  \vspace*{-4mm}
  \label{fig:framework}
  \end{center}
\end{figure}

\begin{algorithm}
\scriptsize
\DontPrintSemicolon
\SetAlgoCaptionSeparator{}
\KwInput{Odometry measurements of robots $i$, $j$ at time $t$: ${\bf{o}}_i^t$, 
${\bf{o}}_j^t$, submap of robots $i$, $j$: ${{\mathop{\rm s}\nolimits} _i^k}$, ${{\mathop{\rm s}\nolimits} _j^l}$, where $1 \le k \le K$, $1 \le l \le L$, and $K$, $L$ are the numbers of corresponding submaps for robot $i$ and robot $j$}
\KwOutput{Constructed pose graph $G$}
 // detect the loop closures and build pose graph \\
 \For{i $ \leftarrow $ 1 \textbf{to} N}{
    \For{k $ \leftarrow $ 1 \textbf{to} K}{
        // obtain feature points of $k^{th}$ submap of robot $i$: ${{\mathop{\rm s}\nolimits} _i^k}$ \\
        \For{j $ \leftarrow $ 1 \textbf{to} N}{
            \For{l $ \leftarrow $ 1 \textbf{to} L}{
                // obtain feature points of $l^{th}$ submap of robot $j$: ${{\mathop{\rm s}\nolimits} _j^l}$ \\
                // calculate the distance between submaps \\
                \If{distance $>$ $\lambda $}
                {
                \textbf{break}
                }
            
                // match ${{\mathop{\rm s}\nolimits} _i^k}$ and   ${{\mathop{\rm s}\nolimits} _j^l}$ \\
                // calculate the relative transformation and consistency confidence score $c_{{i^k},{j^l}}$\\
                \If{$c_{{i^k},{j^l}} > 0.5$}{
                \If{$i \ne j$}{
                Add inter-robot loop closures to $G$
                }
                \If{$i = j$}{
                Add intra-robot loop closures to $G$
                }
                }
            }

        // Add the odometry constraints \\
        $G = G \cup  \textbf{o}_i^t \cup \textbf{o}_j^t$ \\
        }
    }
}
 \KwRet{Pose graph $G$}
 \caption{Pose graph construction for the multi-robot system}
 \label{algo:mapping}
\end{algorithm}
\setlength{\textfloatsep}{8pt}

This section explains one of the main compositions of the proposed DMPF-Explore, which is the multi-robot collaborative mapping. Each robot has a separate mapping module running independently, as illustrated in Fig. \ref{fig:framework}. In this paper, the initial pose of each robot is assumed to be known as a prior. Formally, the robot group consists of $N$ robots, and we denote the pose of robot $i$ ($1 \le i \le N$) at time $t$ as ${\bf{x}}_i^t = [{\bf{R}}_i^t,{\bf{p}}_i^t] \in SE(2)$, in which ${\bf{R}}_i^t$ represents the rotation matrix of the heading $\theta _i^t$ and ${\bf{p}}_i^t = [x_i^t,y_i^t]$ is the 2D position.
The odometry measurement of robot $i$ at time $t$ is described as ${\bf{o}}_i^t \in SE(2)$.
And the relative transformation of odometry measurements between the adjacent timestamps of robot $i$ is denoted as ${\bf{x}}_i^{t - 1,t}$.
In addition, we use ${\overline {\bf{x}} _{{i^n},{j^m}}}$ to represent the submap-based loop closures, which means the relative pose between the $n^{th}$ submap created by robot $i$ and the $m^{th}$ submap created by robot $j$.
To quantify the correlation between two matched submaps, we use ${c_{{i^n},{j^m}}}$ to represent the consistency confidence score.
Due to the pose graph optimization framework being sensitive, a consistency confidence score greater than 0.5 is considered a loop closure. The modules of the DSMC-Map will be explained in detail in the following subsections.

\subsection{Submap construction}
We use a collection of consecutive frames of LiDAR data and corresponding odometry measurements to establish the submap.
It is widely known in the robotics community that odometry measurements cannot satisfy the requirement of LiDAR mapping due to accumulative errors over a long time. Still, they can provide a relatively precise trajectory estimation over a short period of time.
In this paper, we refer to the mapping method proposed in \cite{Karto_SLAM} to create the occupancy grid submap, which performs the correlative scan matching algorithm to optimize consecutive LiDAR scans and build high-accuracy submap.

\subsection{Loop closure detection}
The loop closure detection is a fundamental module for robotic teams to complete multi-robot collaborative mapping tasks in unknown enviornments.
We refer to the method in \cite{SMMR} to calculate the relative transformation and consistency confidence score between two submaps.
Moreover, we perform matching of two submaps if the distance between their odometry pose is less than a certain threshold $\lambda$ to save computational resources.
So, firstly, we utilize the Accelerated-$KAZE$ ($AKAZE$) \cite{AKAZE} local feature descriptor based on the Modified Local Difference Binary (MLDB) algorithm to extract the feature points in the submaps.
This method has been implemented into OpenCV library\cite{opencv}.
The extracted feature points of the submap of robot $i$ at time $t$ are defined as ${{\rm{s}}_i^t}$.
Then, the feature matcher, $AffineBestOf2NearestMatcher$ \cite{Matcher_opencv}, is used to find the best match for each feature point, which also will calculate the relative transformation ${{\overline {\bf{x}}}_{{i^m},{j^n}}}$ and consistency confidence score ${c_{{i^m},{j^n}}}$ between two submaps.  
When $i \ne j$, the estimated relative transformation is considered as inter-robot loop closures.
Otherwise, when $i = j$ and $m \ne n$, the estimated relative transformation will be considered as intra-robot loop closures.

\subsection{Distributed pose graph optimization}
Considering the challenge of computational and communicational resources in the centralized system, this paper uses a DPGO \cite{DPGO} to optimize the trajectories of a group of robots based on LiDAR loop closures and odometry measurements.
DPGO can be modeled as a pose graph $G$ consisting of collections of vertices and edges.
In general, the vertices are denoted as the poses of robots to be estimated, and the edges represent the relative pose estimation between two vertices (odometry or LiDAR loop closures).
After completing the pose graph construction, we can solve that by minimizing the residual error, which can be expressed as

\setlength\lineskip{0pt}
\begin{scriptsize}
\[\begin{array}{l}
\mathop {{\rm{minimize}}}\limits_{{\bf{x}}_i^t} \underbrace {\sum\limits_{i = 1}^N {\bigg(||{\bf{R}}_i^t - {\bf{R}}_i^{t - 1}{\Delta {\bf{R}}}_i^{t - 1,t}||_2^2 + ||{\bf{p}}_i^t - {\bf{p}}_i^{t - 1} - {\bf{R}}_i^t{\Delta {\bf{p}}}_i^{t - 1,t}||_2^2}\bigg) }_{{\rm{\text{Odometry Constraints}}}}\\
\begin{array}{*{20}{c}}
{}&{}&{ + \underbrace {\sum\limits_{i = 1}^N {\sum\limits_{j = 1}^N {\bigg(||{\bf{R}}_j^m - {\bf{R}}_i^n{\overline {\bf{R}}}_{i^n,j^m}||_2^2 + ||{\bf{p}}_j^m - {\bf{p}}_i^n - {\bf{R}}_i^n{\overline {\bf{p}}}_{i^n,j^m}||_2^2\bigg)} } }_{{\rm{\text{Intra-robot and Inter-robot Constraints}}}}}
\end{array} 
\end{array}\]
\end{scriptsize}
\setlength\lineskip{0pt}
\hfill (1)

\noindent where ${\bf{R}}_i^{t - 1} $ and ${\bf{p}}_i^{t - 1}$ represent the rotation and translation components of robot $i$ at time $t-1$. $\Delta{\bf{R}}_i^{t - 1,t}$ and $\Delta{\bf{p}}_i^{t - 1,t}$ are the relative rotation and translation of odometry of robot $i$ from $t-1$ to $t$, respectively. ${{\overline {\bf{R}}}_{i^n,j^m}}$ and ${{\overline {\bf{p}}}_{i^n,j^m}}$ represent the relative rotation and translation of LiDAR loop closures between the poses of robot $i$ at time $n$ and robot $j$ at time $m$.

To solve eq. (1), we utilize the method proposed by \cite{DPGO}, which is based on a sparse semidefinite relaxation of the pose graph optimization problem and is able to confront the challenge from large-scale environments. 
Afterward, the estimated poses and corresponding LiDAR data are used to construct the global map.
The process of constructing the pose graph is shown in Algorithm \ref{algo:mapping}.

\section{MWF-CN exploration strategy}
\label{section:exploration}

\begin{figure}
  \begin{center}
  \includegraphics[width=\linewidth]{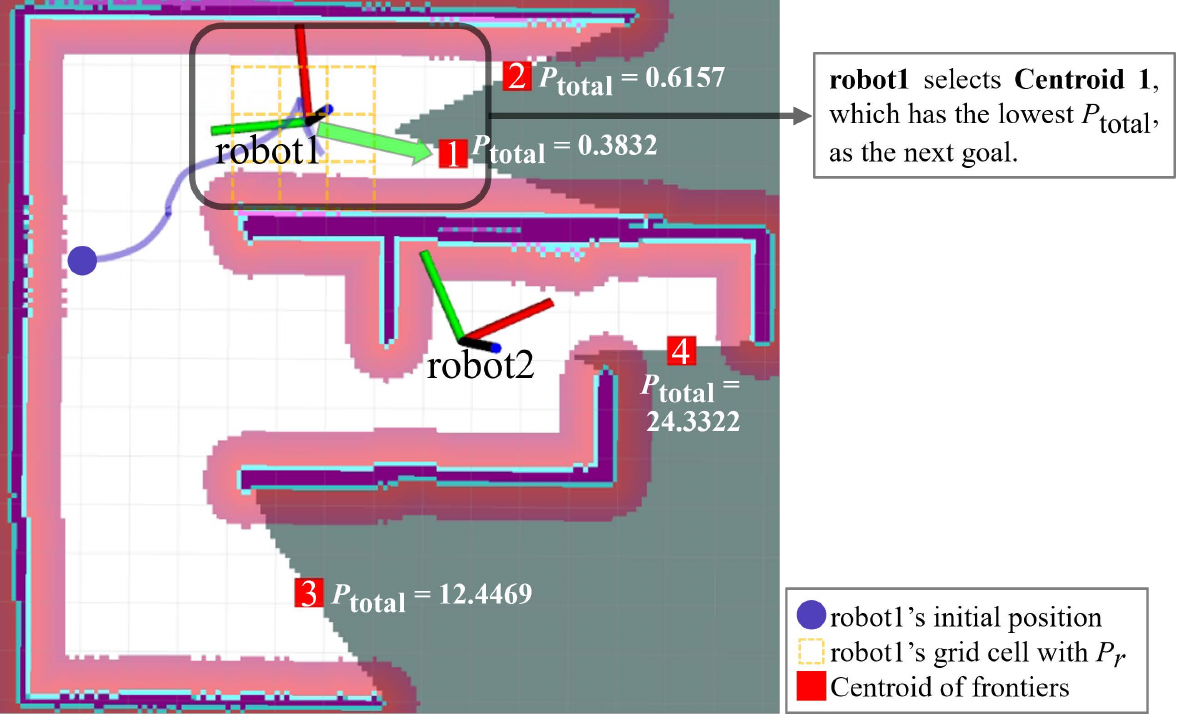}
  \caption{Graphical representation of the exploration by MWF-CN from robot1's point of view. $P_\text{total}$ is the total potential calculated by eq. (2), $P_r$ is the repulsive potential calculated by eq. (9), and the green arrow is the robot1's desired direction to reach the goal at the centroid with the lowest $P_\text{total}$.}
  \vspace*{-4mm}
  \label{fig:MWF-CN}
  \end{center}
\end{figure}

This section describes the details of the proposed exploration strategy for multiple collaborative robots, MWF-CN. First of all, the detected frontiers are clustered by the continuity-based algorithm \cite{SMMR}, and the centroid of each cluster will represent all the frontiers in that corresponding cluster. Then, the total potential, which consists of the attractive potential of frontiers and the repulsive potential of robots, will be calculated and assigned to the centroids. Each robot will first try to reach the centroid with the lowest potential value, as shown in Fig. \ref{fig:MWF-CN}. The total potential of robot $i$ for centroid $q(x_q,y_q)$ at point $p(x_p,y_p)$ is defined as
\begin{equation}\setcounter{equation}{2}
    P_{\text{total}} (i,p,q) = P_a(p,q) + P_r(i,p)\text{,}
    \label{eq:P_total}
\end{equation}
where $P_a(p,q)$ is the attractive potential of the frontier centroid $q$ at point $p$ and $P_r(i,p)$ is the repulsive potential of robot $i$ at point $p$. The potential modeling attractive effect is used for leading robots to explore the unknown area of environments. On the other hand, repulsiveness oversees in-between robot interactions. The overall calculation process is presented in Algorithm \ref{algo:strategy}, and the compositions of the potential field will be explained further below in the following subsections.

\begin{algorithm}
\scriptsize
\DontPrintSemicolon
\SetAlgoCaptionSeparator{}
  \KwInput{Current cluster that centroid $q$ belongs to, where $C_q$ is the size of the cluster, orthogonal and diagonal neighbors of the centroid $q$, MWF distances between point $p$ and centroid $q$: $d^*(p,q)$, distances between robot $i$ and point $p$: $d(i,p)$, number of robots $N$}
  \KwOutput{Total potential of robot $i$ for centroid $q$ at point $p$: $P_\text{total}$} 
  \For{l $ \leftarrow $ 1 \textbf{to} 8 (number of neighbors)}
  {
    $P_a = 0$, $P_r = 0$
    
    // calculate the attractive potential
    
    \For{m $ \leftarrow $ 1 \textbf{to} $C_q$}{
        $P_a -= k_a C_q / d^* (p,q)$
        
        }
        
    // calculate the repulsive potential
    
    \For{n $ \leftarrow $ 1 \textbf{to} N}{
        \If{$d(i,p) < d_s$} {
        $P_r += k_r \text{exp}\Big(\frac{d(i,p) - d_s}{\sigma_r}\Big) + \chi^p_i (\alpha, \sigma_d)$

        }
    }
    
    // calculating the total potential
    
    $P_\text{total} = P_a + P_r$
    
    }
\KwRet{$P_\text{total}$}
\caption{Attractive and repulsive potentials calculation}
\label{algo:strategy}
\end{algorithm}

\subsection{Attractive potential of frontiers}

The attractive potential of frontiers attracts the robots to appropriate goals based on the potential value. As previously stated, the frontiers detected from the merged map are clustered, and their centroids' potentials will be calculated. The attractive potential of centroid $q$ at point $p$ is defined as

\begin{equation}
    P_a (p,q) = -\frac{k_a C_q}{d^* (p,q)} \text{,}
    \label{eq:P_a}
\end{equation}
where $k_a$ is a scaling constant, $C_q$ is the total number of frontiers in the cluster that centroid $q$ belongs to, and the leading role of this potential is the MWF distance from point $p$ to centroid $q$ denoted by $d^* (p,q)$. 

\begin{figure*}
     \centering
     \begin{subfigure}[b]{0.49\linewidth}
         \centering
         \includegraphics[width=\linewidth]{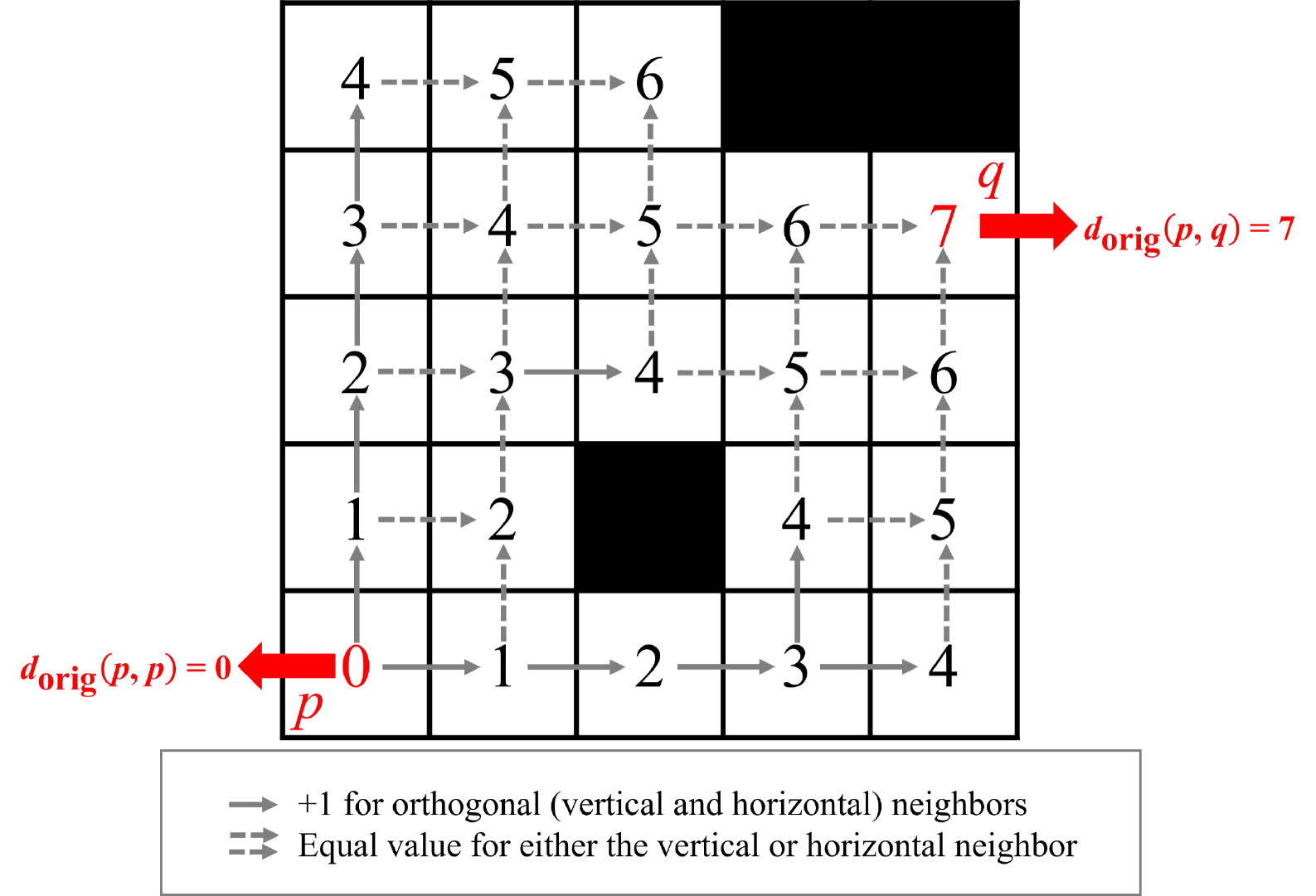}
         \caption{Original wave-front distance ($d_\text{orig}$)}
         \label{fig:Orig-WF-example}
     \end{subfigure}
     \begin{subfigure}[b]{0.49\linewidth}
         \centering
         \includegraphics[width=\linewidth]{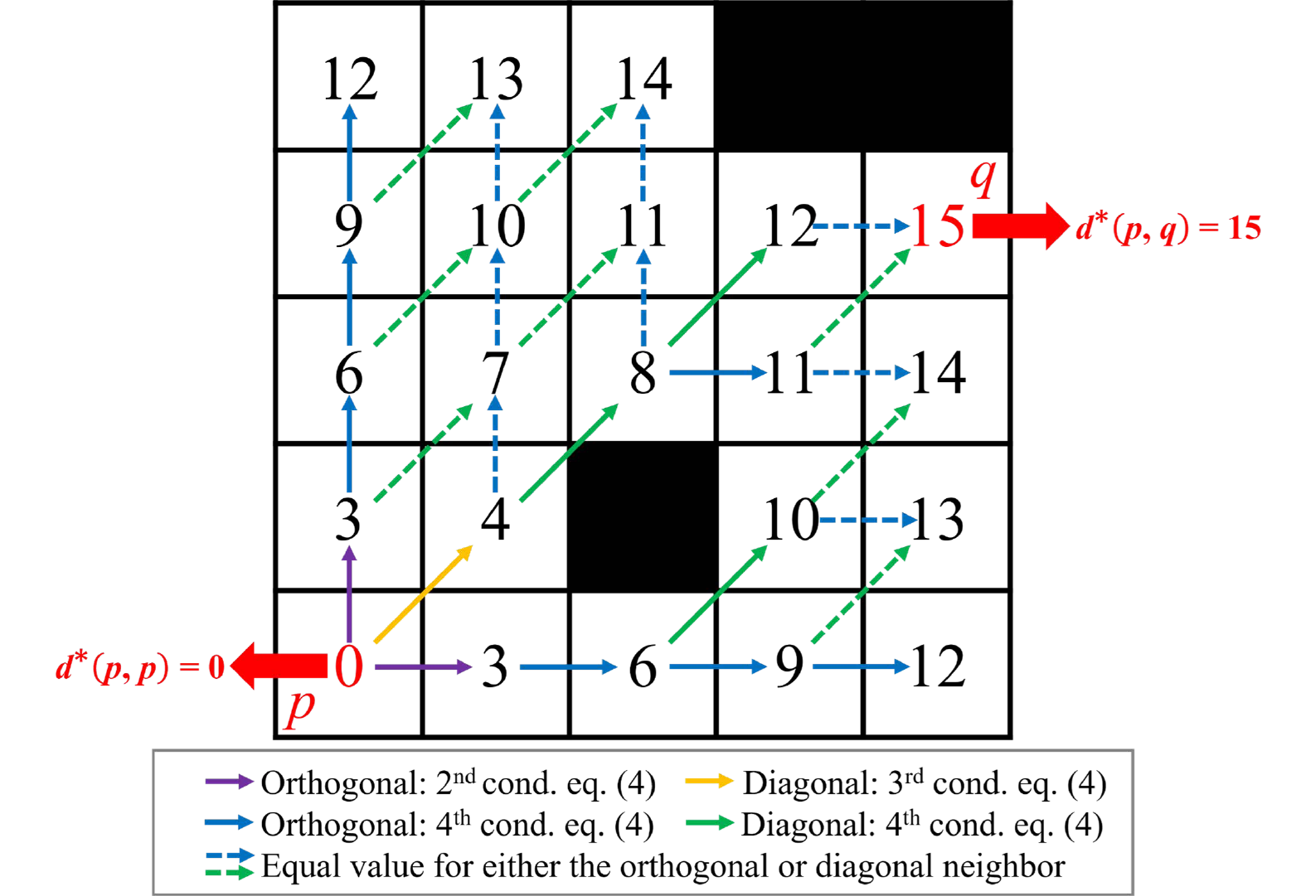}
         \caption{MWF distance ($d^*$)}
         \label{fig:MWF-example}
     \end{subfigure}
        \caption{Examples of the calculation by different types of wave-front distances}
        \label{fig:WF-examples}
\end{figure*}

A similar distance form was once introduced in \cite{MWFRef}, and there were some studies regarding the comparison between the performance by different types of distances \cite{WFvsA*}, \cite{WFvsMWF}. However, the previous ones were not utilized as a part of the potential. Moreover, the mathematical expression of this distance function has never been constructed explicitly before. Hence, for the purposes of clarification and further development, here we present the MWF distance $d^* (p,q)$ from point $p$ to centroid $q$ newly defined as

\begin{equation}
    d^* (p,q) =
    \left\{
    \begin{array}{ll}
        0 & q = p\text{,} \\
        3 & q \in \mathcal{N}_{orth}(p) \setminus \mathcal{O}\text{,} \\
        4 & q \in \mathcal{N}_{diag}(p) \setminus \mathcal{O}\text{,} \\
        \min(\mathcal{D} (k,q)) & \text{otherwise,}
    \end{array} 
    \right. 
\end{equation}
where $\mathcal{O}$ is the set of obstacles. The neighborhood of $p$ is
\begin{equation}
    \mathcal{N}(p) = (\{x_p - 1, x_p, x_p + 1\} \times \{y_p - 1, y_p, y_p + 1\}) \setminus \{(x_p, y_p)\} \text{.}
\end{equation}
The diagonal and orthogonal neighborhoods of point $p$ are
\begin{align}
    \mathcal{N}_{diag}(p) &= \{x_p - 1, x_p + 1\} \times \{y_p - 1, y_p + 1\}\text{,} \\
    \mathcal{N}_{orth}(p) &= \mathcal{N}(p) \setminus \mathcal{N}_{diag}(p) \text{.}
\end{align}
And the set of MWF distances from point $q$'s neighborhood to $q$ is defined as
\begin{align}
    \mathcal{D} (k, q) = &\big\{d^* (k,q) + 3 \mid k \in \mathcal{N}_{orth}(q) \setminus \mathcal{O}\big\} \notag
    \\ &\cup \big\{d^* (k,q) + 4 \mid k \in \mathcal{N}_{diag}(q) \setminus \mathcal{O}\big\}\text{.}
\end{align}
Note that the neighborhoods exist only on the explored areas and frontiers. The calculation process starts at point $p$, where $d^*(p,p) = 0$, and continues recursively until reaching centroid $q$, as shown in Fig. \ref{fig:MWF-example}.

Since some computational time is eliminated due to obstacles being considered in the MWF distance, a separate potential for enhancing collision avoidance is not needed. Furthermore, the MWF distance is formulated based on 8-sector connectivity, including 4 orthogonal and 4 diagonal sectors, which is different from the original version used in the MMPF algorithm \cite{SMMR}, in which the wave-front distance was based on the 4-sector orthogonal neighborhood with the same $+1$ distance weights only, as shown in Fig. \ref{fig:Orig-WF-example}. Since the accessible neighborhood here is extended from 4 to 8 sectors, the frontiers can be detected more, even in the small areas. Also, different distance values $+3$ and $+4$ are assigned to the orthogonal and diagonal neighbors, respectively. By utilizing the presented MWF distance, this work offers a novel approach to the more extensive frontier neighborhoods for the potential field, which also reduces the selection and local optima problems from equal potential values.

\subsection{Repulsive potential of robots}

Turning now to another significant potential, each robot's repulsive potential is assigned to separate the robots when they are too close to each other and tend to explore the same area. Here the repulsive potential of robot $i$ at point $p$ is defined as
\begin{equation}
    P_r(i,p) =
    \left\{
    \begin{array}{ll}
        k_r e^{\frac{d(i,p) - d_s}{\sigma_r}} + \chi^p_i(\alpha,\sigma_d) &d(i,p) < d_s\\
        0 & d(i,p) \geq d_s\\
    \end{array} 
    \right.
\label{eq:P_r}
\end{equation}
where $d_s$ is the sensor range, $d(i,p)$ is the distance between robot $i$ and point $p$, $k_r$ is the scaling constant, $\sigma_r$ is the relaxation distance, and $\chi_i^p(\alpha,\sigma_d)$ is the colored noise term for robot $i$ at point $p$ with the noise color $\alpha$ and variance $\sigma_d$.

We choose the first term of the potential to be exponentially decreasing with a relaxation distance $\sigma_r$ in order to make the potential not as strict as linear functions, which is used in [10]. The chosen exponential potential leads to more flexible exploration manners, which are essential for real-world situations where robots should not be too “self-preserved,” e.g., conflagration, earthquake, mass shootings, and other urgencies. The distance $\sigma_r$ also acts as the desired distance for the robots to be far from each other.  

Moreover, the colored noise \cite{Noise1995,Noise2011}, also known as the $1/f^\alpha$ noise, with noise color $\alpha$ and variance $\sigma_d$ is added to the potential as the fluctuation, in which to the best of our knowledge, no other works have applied it for the robot exploration before. The name originates from the noise spectra that grow as $1/f^\alpha$, where $f$ is the cyclic frequency, and $\alpha$ is a real number in the interval $[0,2]$ indicated the color of the noise. The white noise ($\alpha = 0$) is uncorrelated; each new value is independent of the previous ones. In contrast, the brown noise ($\alpha = 2$) is strongly correlated since it is an integral of the white noise. And the last one, the pink noise ($\alpha = 1$), is between the two noises mentioned earlier. Colored noises have been studied in \cite{Noise1, Noise2} that they can influence the dynamics and trigger complex patterns. Hence, they are adopted in various stochastic models \cite{Noise3, Noise4, Noise5}. This study provides a novel approach for applying color noises in multi-robot exploration. Since different $\alpha$ and variance $\sigma_d$ produce distinct patterns of noise sequences, the main motive for this development is to diversify the potential values by the choosable parameters for overcoming different environmental conditions. The Finite Impulse Response (FIR) method is utilized for generating the noise sequences based on
\begin{equation}
    \Delta \chi_k = \chi_{k+1} - \chi_k = w_{k+1} + \bigg( \frac{\alpha}{2} \bigg) \sum_{m=0}^{k-1} \frac{h_m}{m+1}w_{k-m} \text{, }
\end{equation}
where $w_k$ the sequence of the Gaussian white noise with mean 0 and variance $\sigma_d$. The $h_k$ is the pulse response defined as
\begin{equation}
    h_k = \frac{\Gamma(\alpha/2 + k)}{k!\Gamma(\alpha/2)} \text{,}
\end{equation}
which can be simply computed by the recursion
\begin{align}
    h_0 &= 1 \notag \\
    h_k &= \bigg( \frac{\alpha}{2} + k - 1 \bigg)\frac{h_{k-1}}{k}\text{.}
\end{align}

\begin{figure*}[H]
  \begin{center}
  \includegraphics[width=\linewidth]{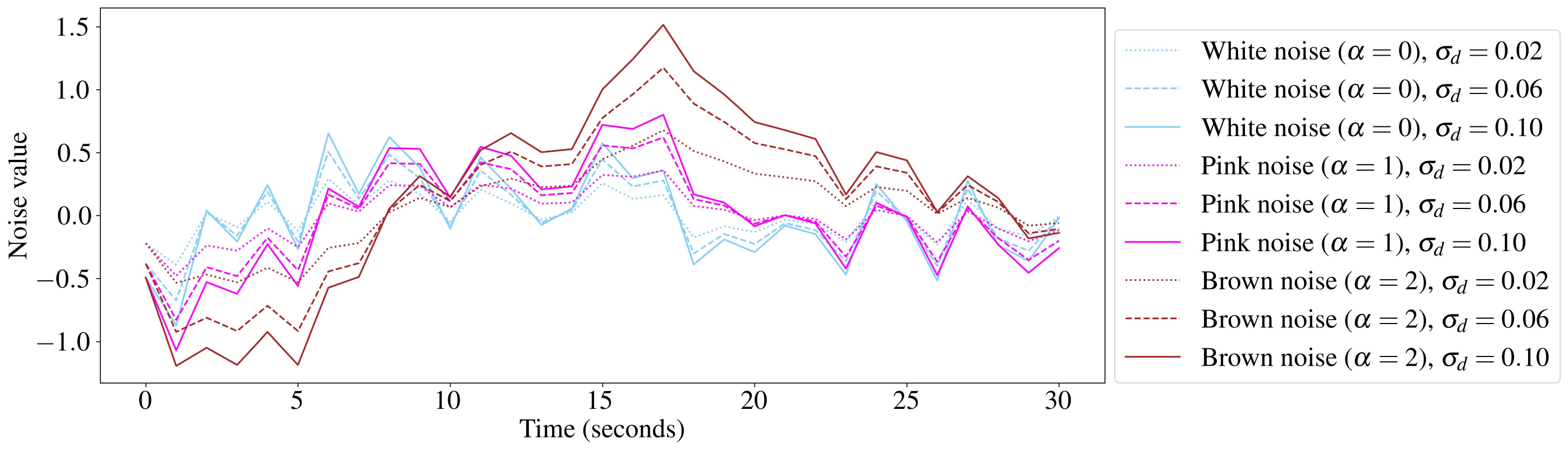}
  \caption{Example of noises generated by different noise colors $\alpha$ and variances $\sigma_d$ with the sampling rate at 1 Hz}
  \label{fig:noises}
  \vspace*{-6mm}
  \end{center}
\end{figure*}

For the repulsive potential, we would like to have diverse potential values to match the conditions of distinct environments. According to different impacts, we select three parameters in each scenario: $\sigma_r$, $\alpha$, $\sigma_d$. Firstly, $\sigma_r$ influences the desirable distance between robots, i.e., a lower value of $\sigma_r$ allows the robots to be closer to each other. For $\alpha$ and $\sigma_d$, these parameters are for the noise. As we can observe from Fig. \ref{fig:noises}, different noise colors $\alpha$ produce non-identical behaviors of noise values, and the noise variance $\sigma_d$ plays a part in the intensity of the noise manners. So, it is likely that we should choose the high $\alpha$ and balanced $\sigma_d$ if the environment is complex in order to deal with the potential values of several frontier centroids. For the best performance, further investigation should be done to find suitable parameter values.

\section{Simulation}
\label{section:simulation}

\begin{figure}
     \centering
     \begin{subfigure}[b]{0.4\linewidth}
         \centering
         \includegraphics[width=\linewidth]{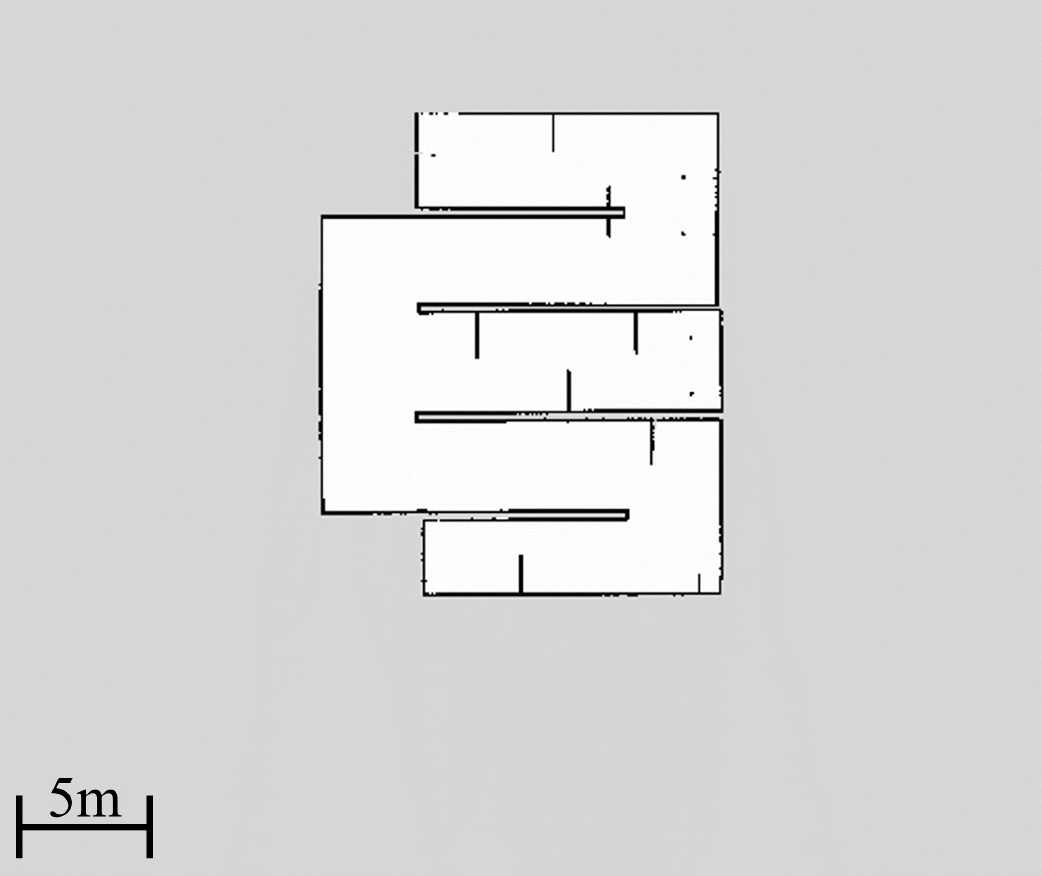}
         \caption{Map 1}
         \label{fig:map1}
     \end{subfigure}
     \begin{subfigure}[b]{0.4\linewidth}
         \centering
         \includegraphics[width=\linewidth]{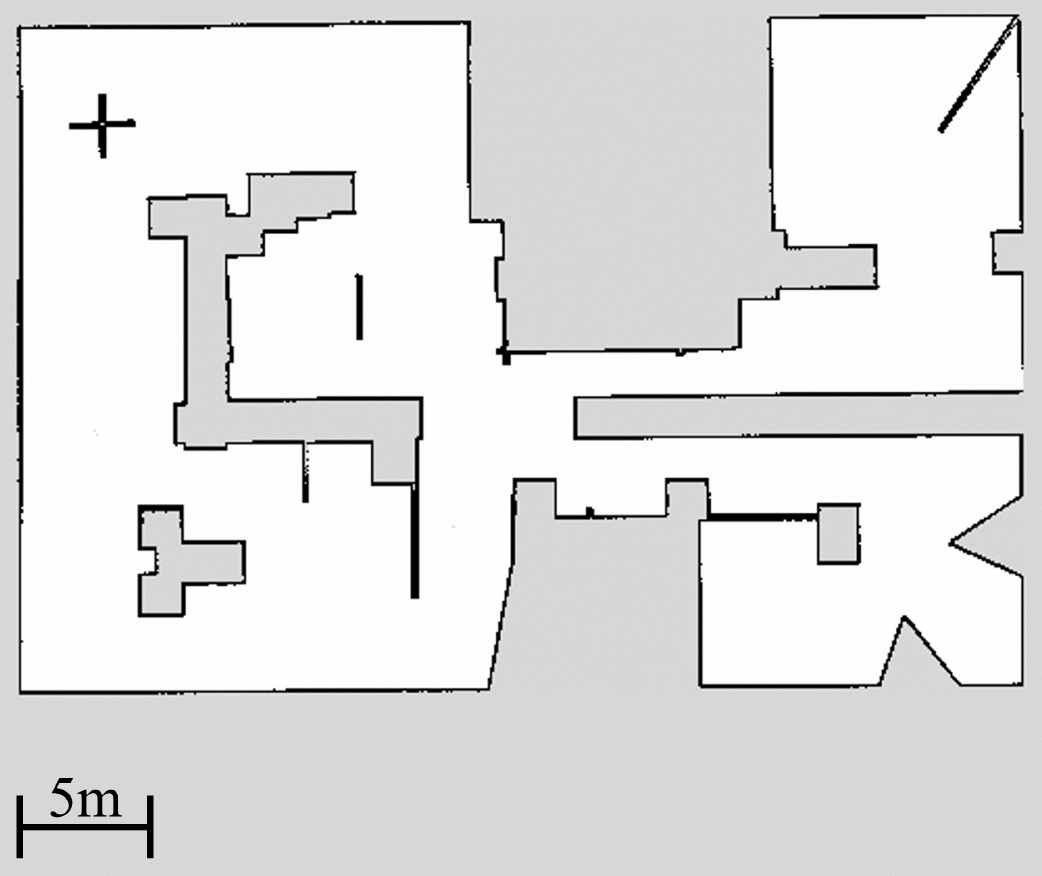}
         \caption{Map 2}
         \label{fig:map2}
     \end{subfigure}
        \caption{Environments for simulation}
        \label{fig:maps}
\end{figure}

The simulation of the DMPF-Explore is described in this section. In the first subsection, we describe the metrics used for evaluating exploration methods in this paper. Then, in the second subsection, we focus on finding the best-performing parameters for the MWF-CN exploration strategy that leads to the fastest exploration time. For the third subsection, we conduct the simulation of different exploration methods and compare the performance based on the evaluation metrics. And finally, in the fourth subsection, we look into the computation aspects of each method.

All simulations are conducted using ROS Melodic with Ubuntu 18.04 on a Desktop PC with Xeon(R) CPU E5-1680 v3 @ 3.20GHz × 16 and 31.3 GB RAM. We utilize TurtleBots with 2D LiDAR (7m maximum scanning range) for each as simulation robots. The mapping is done using the DSMC-Map and the mapping method from the benchmark \cite{Explore-Bench}, which consists of Cartographer \cite{cartographer} and its map merger. The TEB planner is used for navigation. We conduct two-robot and three-robot exploration in two Gazebo world environments, in which Map 1 from \cite{TM-RRT} is 120$\text{m}^2$, and Map 2 from \cite{SMMR} is 391$\text{m}^2$, as shown in Fig. \ref{fig:maps}. In addition, the exploration starting points are at the middle left of both maps, and the robots are placed next to each other in the same position.

\subsection{Evaluation metrics}
\label{subsection:metrics}

\begin{figure*}
     \centering
     \begin{subfigure}[b]{0.4\linewidth}
         \centering
         \includegraphics[width=\linewidth]{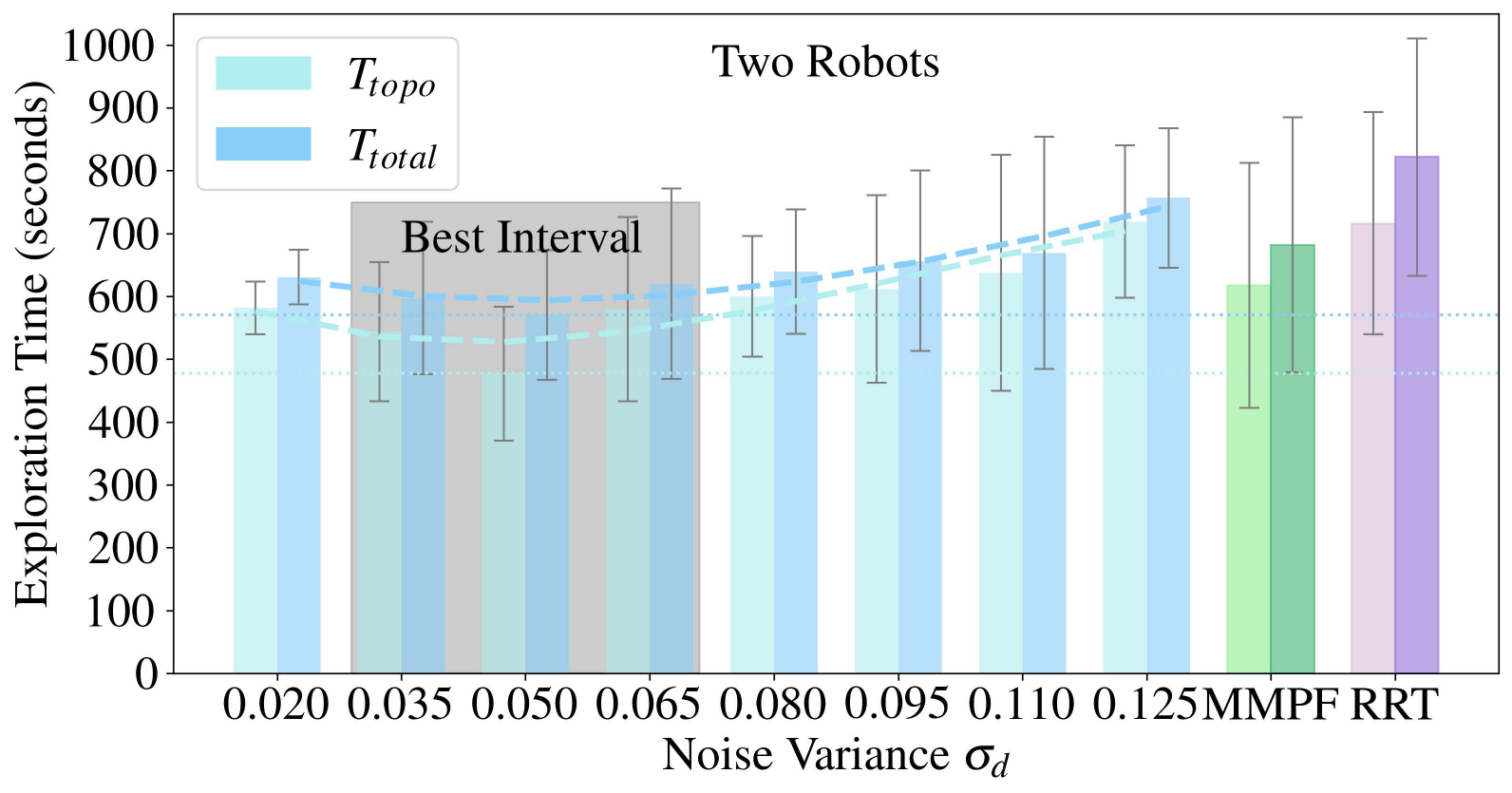}
         \caption{Two robots: White noise ($\alpha=0$)}
         \label{fig:two_alpha=0}
     \end{subfigure}
     \begin{subfigure}[b]{0.4\linewidth}
         \centering
         \includegraphics[width=\linewidth]{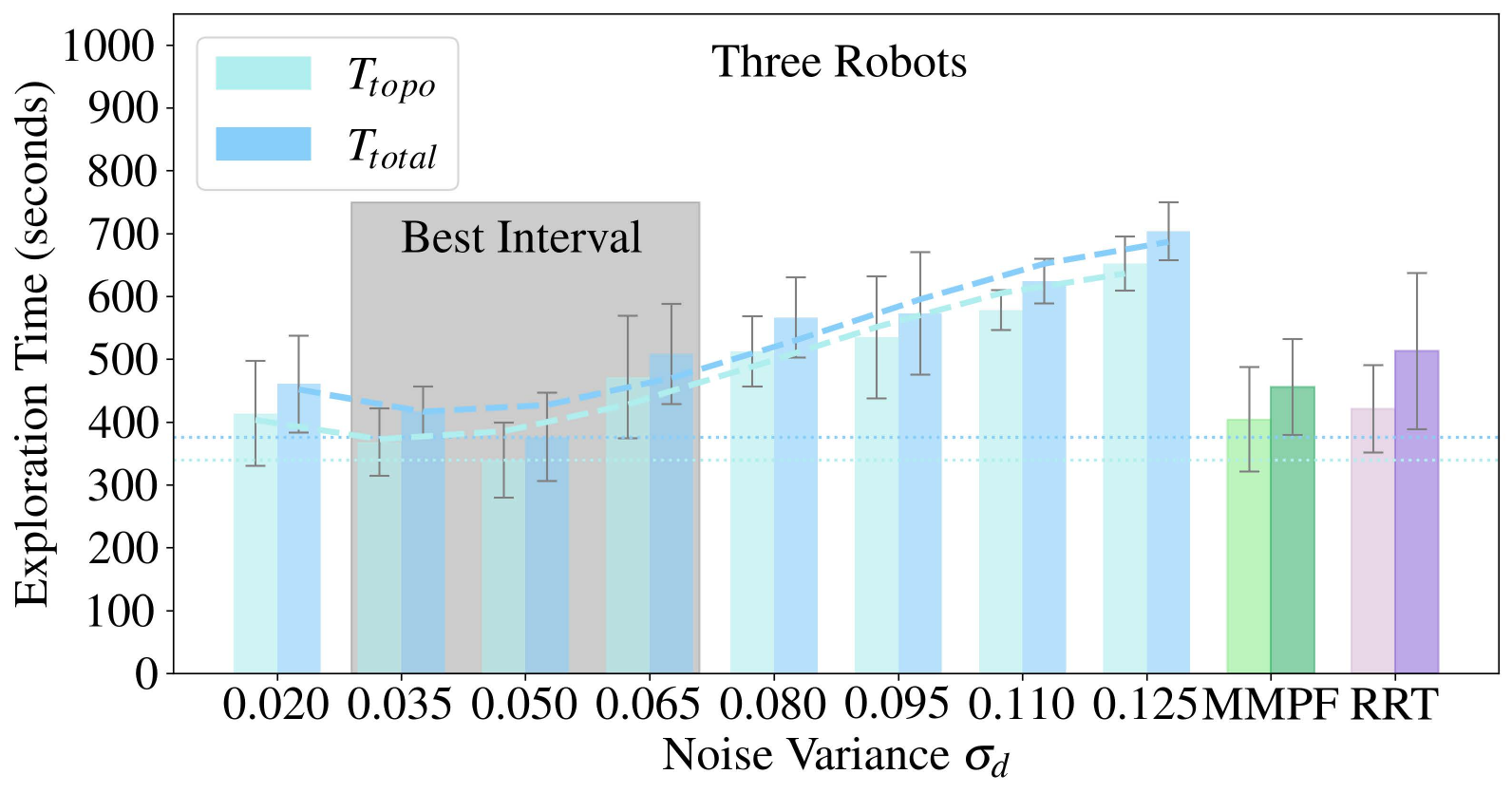}
         \caption{Three robots: White noise ($\alpha=0$)}
         \label{fig:three_alpha=0}
     \end{subfigure}
     \begin{subfigure}[b]{0.4\linewidth}
         \centering
         \includegraphics[width=\linewidth]{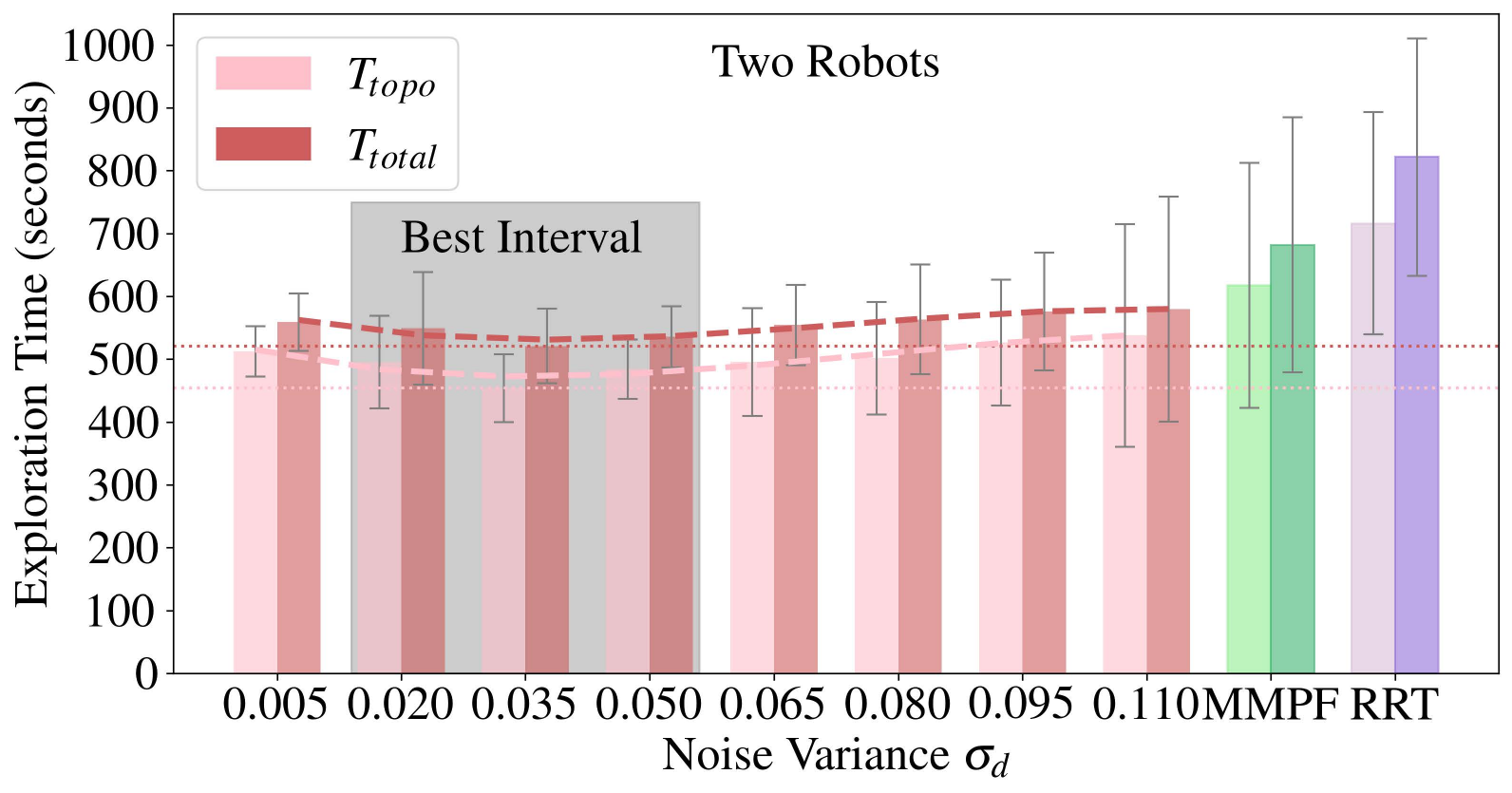}
         \caption{Two robots: Pink noise ($\alpha=1$)}
         \label{fig:two_alpha=1}
     \end{subfigure}
     \begin{subfigure}[b]{0.4\linewidth}
         \centering
         \includegraphics[width=\linewidth]{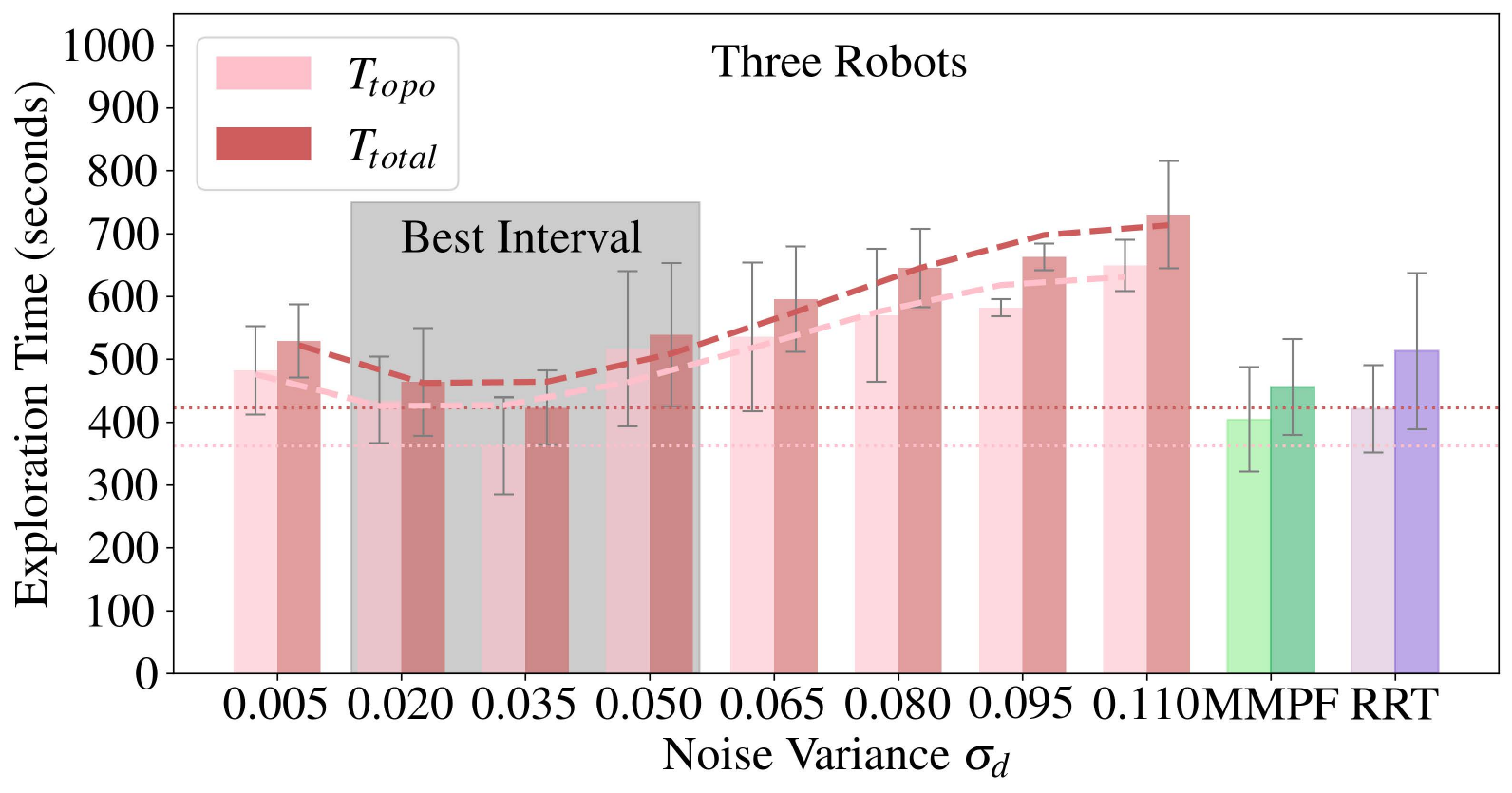}
         \caption{Three robots: Pink noise ($\alpha=1$)}
         \label{fig:three_alpha=1}
     \end{subfigure}
     \begin{subfigure}[b]{0.4\linewidth}
         \centering
         \includegraphics[width=\linewidth]{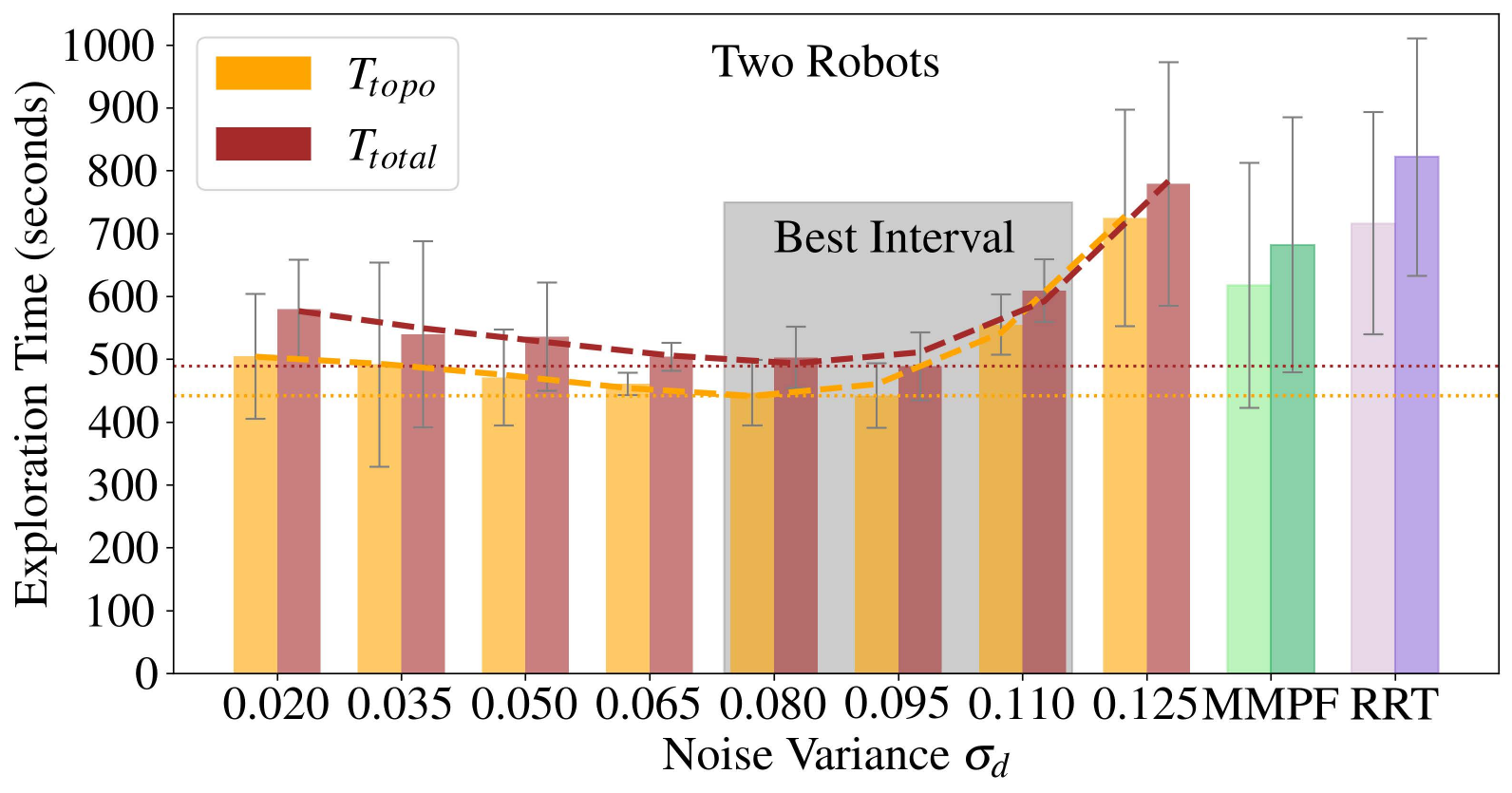}
         \caption{Two robots: Brown noise ($\alpha=2$)}
         \label{fig:two_alpha=2}
     \end{subfigure}
     \begin{subfigure}[b]{0.4\linewidth}
         \centering
         \includegraphics[width=\linewidth]{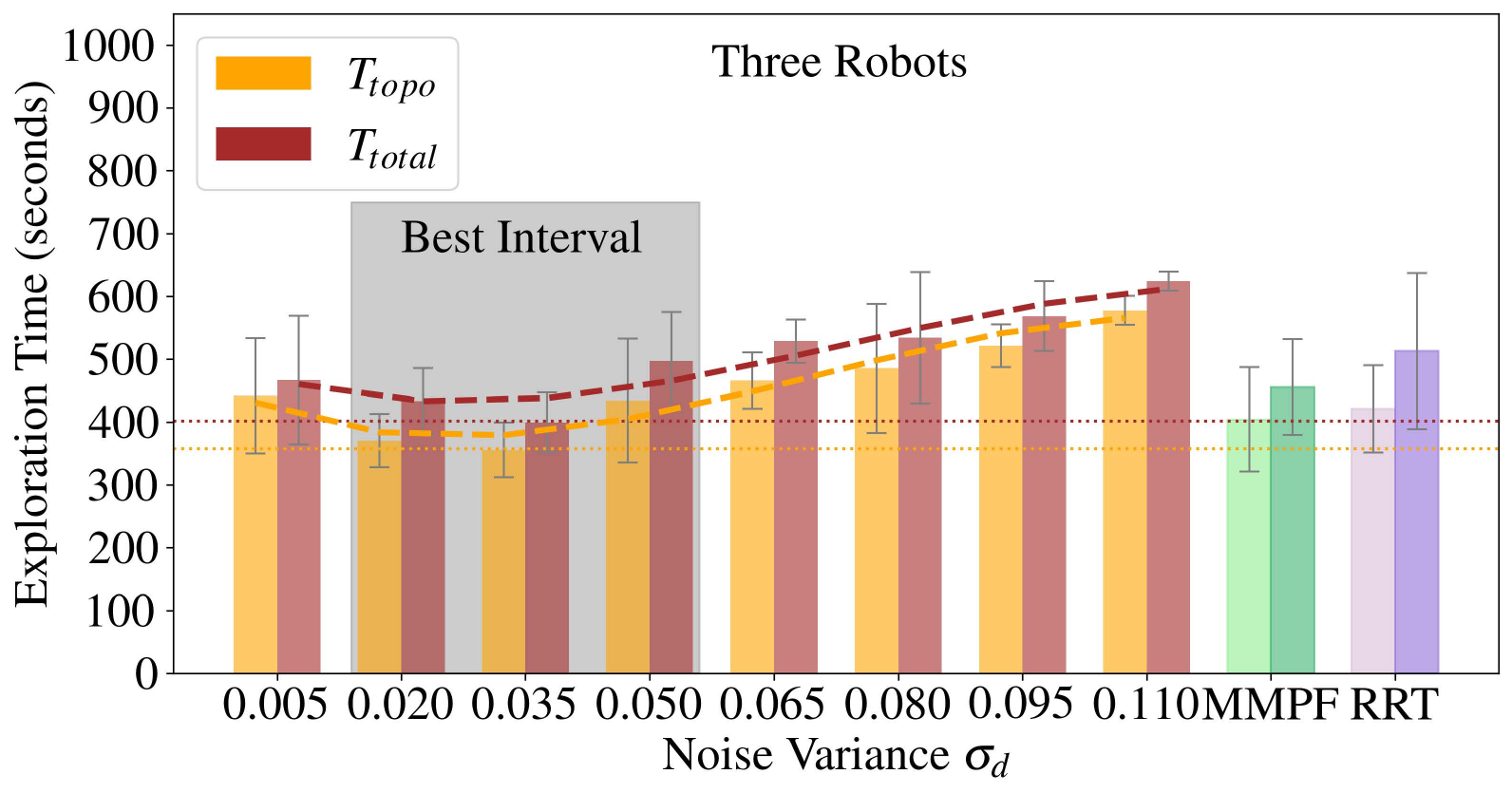}
         \caption{Three robots: Brown noise ($\alpha=2$)}
         \label{fig:three_alpha=2}
     \end{subfigure}
        \setlength{\belowcaptionskip}{-5pt}
        \caption{Simulation results of the MWF-CN (the proposed exploration strategy) with different noise colors $\alpha$ and noise variances $\sigma_d$, MMPF, and RRT against the exploration times $T_\text{topo}$ and $T_\text{total}$}
        \label{fig:optimal}
        \vspace*{-2mm}
\end{figure*}

We follow the metrics in \cite{Explore-Bench} and \cite{SSIM} to evaluate the performance of the proposed approach. These metrics offer an effective way to assess both exploration efficiency and collaboration ability for multiple robots. Hence, they can reflect and differentiate the robustness between each exploration method. The mentioned metrics will be explained below:

\begin{enumerate}
\item \textbf{Topological time ($T_\text{topo}$): }
The topological time is the time robots spend exploring 90\% of the environmental area, generally the overall structure and connectivity. 

\item \textbf{Total time ($T_\text{total}$): }
The total time is the time robots spend exploring 99\% of the environmental area. The reason for not using 100\% is that by the nature of realistic situations, there can be slight errors in the map used for evaluation.

\item \textbf{Standard deviation of the independent exploration areas ($\sigma_\text{ind}$): }
This metric is used for quantitatively evaluating the workload between multiple robots, which is essential for the overall system. 

The $\sigma_\text{ind}$ is defined as follows:
\begin{equation}
    \sigma_\text{ind} = \sqrt{\frac{\sum_{i=1}^N (S_i - \Bar{S})^2 }{N}} \text{,}
\end{equation}
where $N$ is the number of robots, $S_i$, $1 \le i \le N$, is the area explored by the $i$-th robot when finished the exploration, and $\Bar{S}$ is defined as
\begin{equation}
    \Bar{S} = \frac{\sum_{i=1}^N S_i}{N} \text{.}
\end{equation}

\item \textbf{Ratio of the overlapping area explored by multiple robots ($r_\text{overlap}$): }
This metric can verify quantitatively if the robots are efficiently assigned the goals to explore in separate areas.  
The $r_\text{overlap}$ is defined as
\begin{equation}
    r_\text{overlap} = \frac{S_\text{overlap}}{S_\text{total}}  = \frac{\sum_{i=1}^N (S_i - S_\text{total})}{S_\text{total}} \text{,}
\end{equation}
where $S_\text{overlap}$, $S_\text{total}$ are the overlapping and the total areas.

\item \textbf{Success rate ($R_\text{success}$): }
We have this additional metric for the real-world scenarios by considering the exploration to be failed if the robots get ``stuck''. In this paper, we will define it as when the robots' positions do not change for two minutes. 
The success rate $R_\text{success}$ can be expressed as
\begin{equation}
    R_\text{success} = \frac{N_\text{success}}{N_\text{total}} \times 100 \text{,}
\end{equation}
where $N_\text{success}$ is the number of successful exploration rounds and $N_\text{total}$ is the total number of exploration rounds.

\item \textbf{Map's structural similarity index measure ($\text{SSIM}_\text{map}$): }
This metric is additionally used to evaluate the quality of the resulting maps of real-world experiments based on their structural similarity compared to the reference maps. The $\text{SSIM}_\text{map}$ is defined as follows:
\begin{equation}
    \text{SSIM}_\text{map} = \frac{(2 \mu_\text{ref} \mu_\text{res} + c_1)(2\sigma_{\text{ref},\text{res}} + c_2)}{(\mu_\text{ref}^2 + \mu_\text{res}^2 + c_1)(\sigma_\text{ref}^2 + \sigma_\text{res}^2 + c_2)} \text{,}
\end{equation}
where $\mu_\text{ref}$, $\mu_\text{res}$ are the pixel sample means of the reference and resulting maps. $\mu_\text{ref}^2$, $\mu_\text{res}^2$ are the variances of the reference and resulting maps. $\sigma_{\text{ref},\text{res}}$ is the covariance of the reference and resulting maps. And $c_1$, $c_2$ are the variables for stabilizing the division with a weak denominator. Note that the map obtained from the keyboard teleoperation of one robot is used as the reference map for each scenario, and for convenience, we compute the $\text{SSIM}_\text{map}$ by utilizing this metric from the \textit{scikit-image} library \cite{scikit-image}.

\end{enumerate}

\begin{figure*}[]
     \centering
     \begin{subfigure}[b]{0.4\linewidth}
         \centering
         \includegraphics[width=\linewidth]{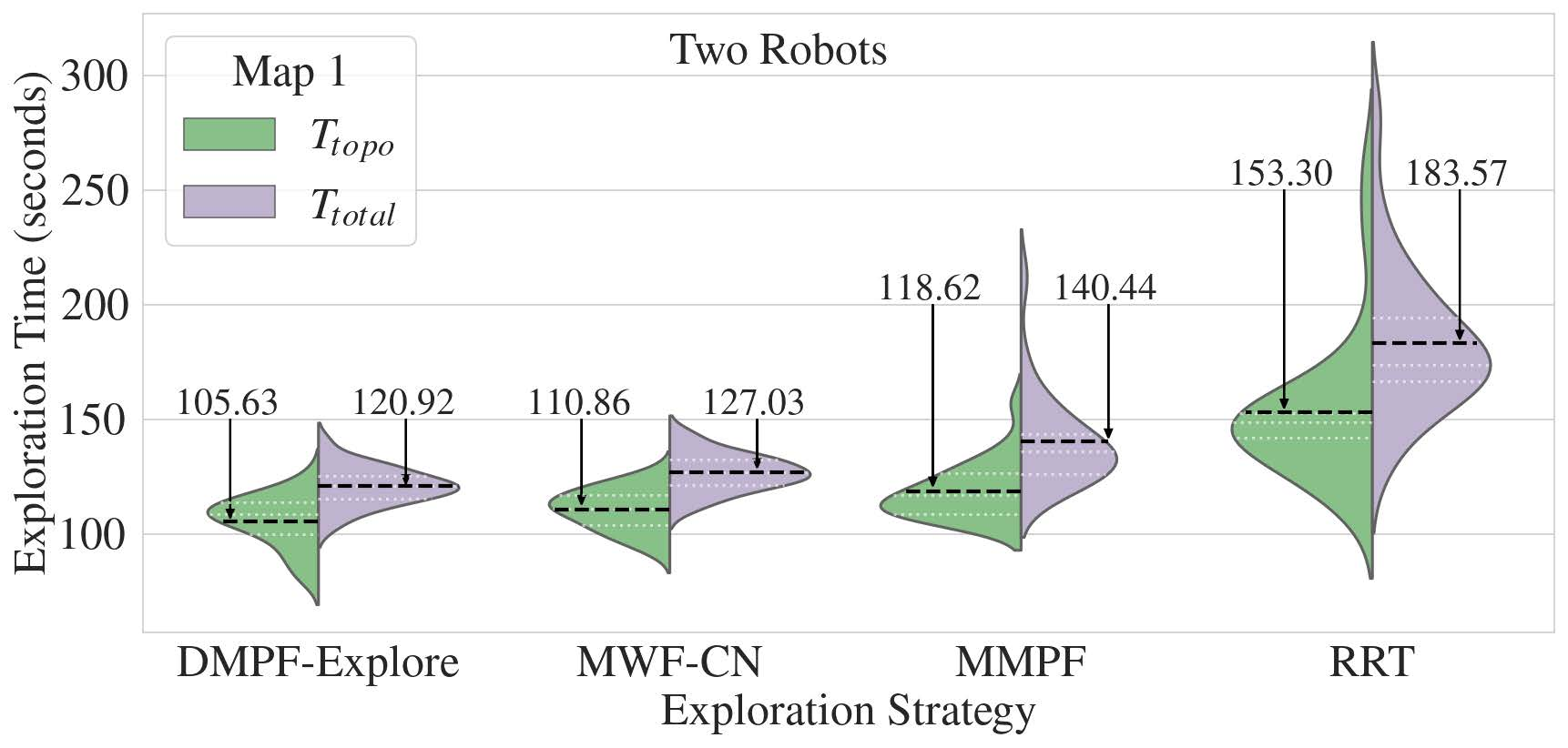}
         \caption{Two robots: $T_\text{topo}$ and  $T_\text{total}$ of Map 1}
         \label{fig:two_sim_map1_time}
     \end{subfigure}
     \begin{subfigure}[b]{0.4\linewidth}
         \centering
         \includegraphics[width=\linewidth]{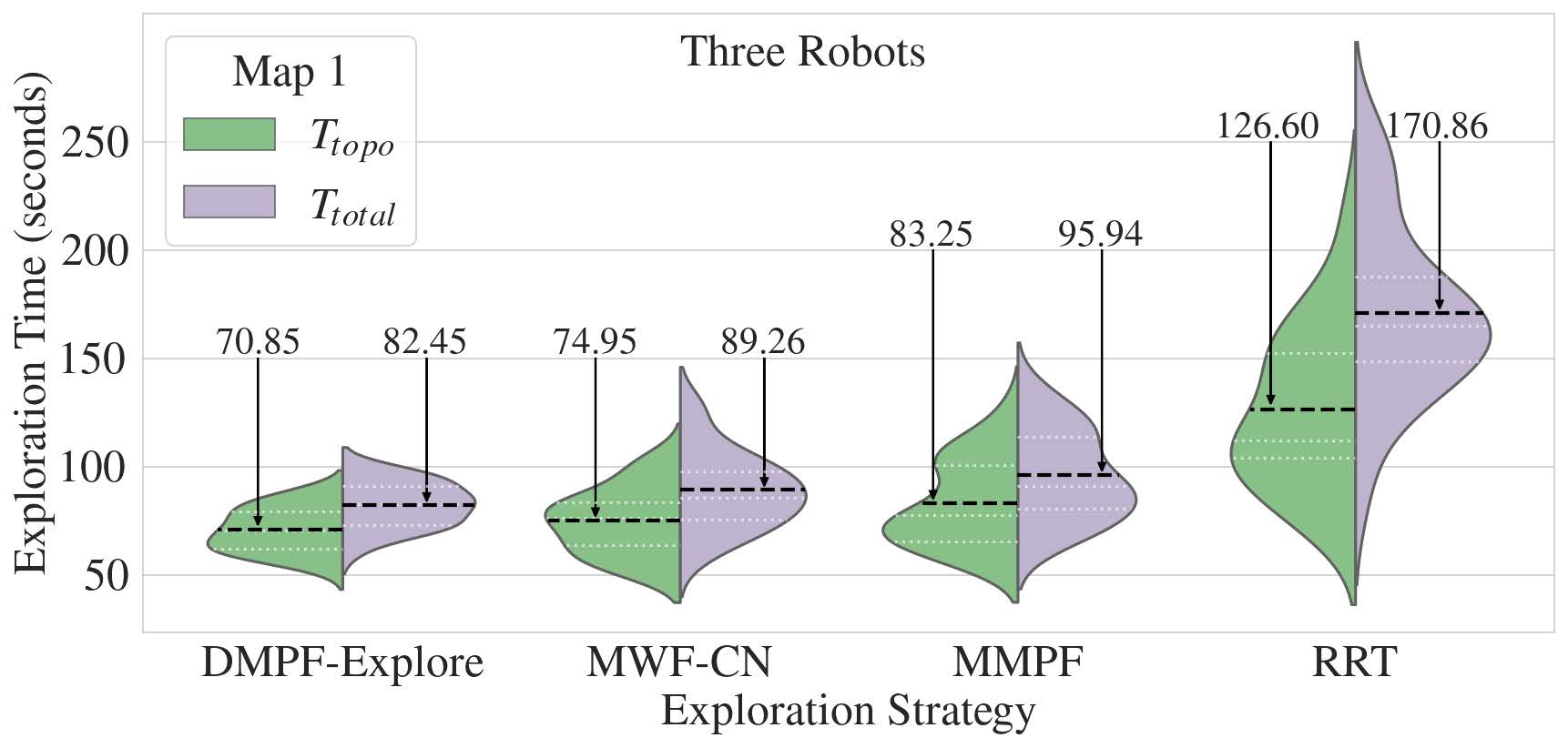}
         \caption{Three robots: $T_\text{topo}$ and  $T_\text{total}$ of Map 1}
         \label{fig:three_sim_map1_time}
     \end{subfigure}
     \begin{subfigure}[b]{0.4\linewidth}
         \centering
         \includegraphics[width=\linewidth]{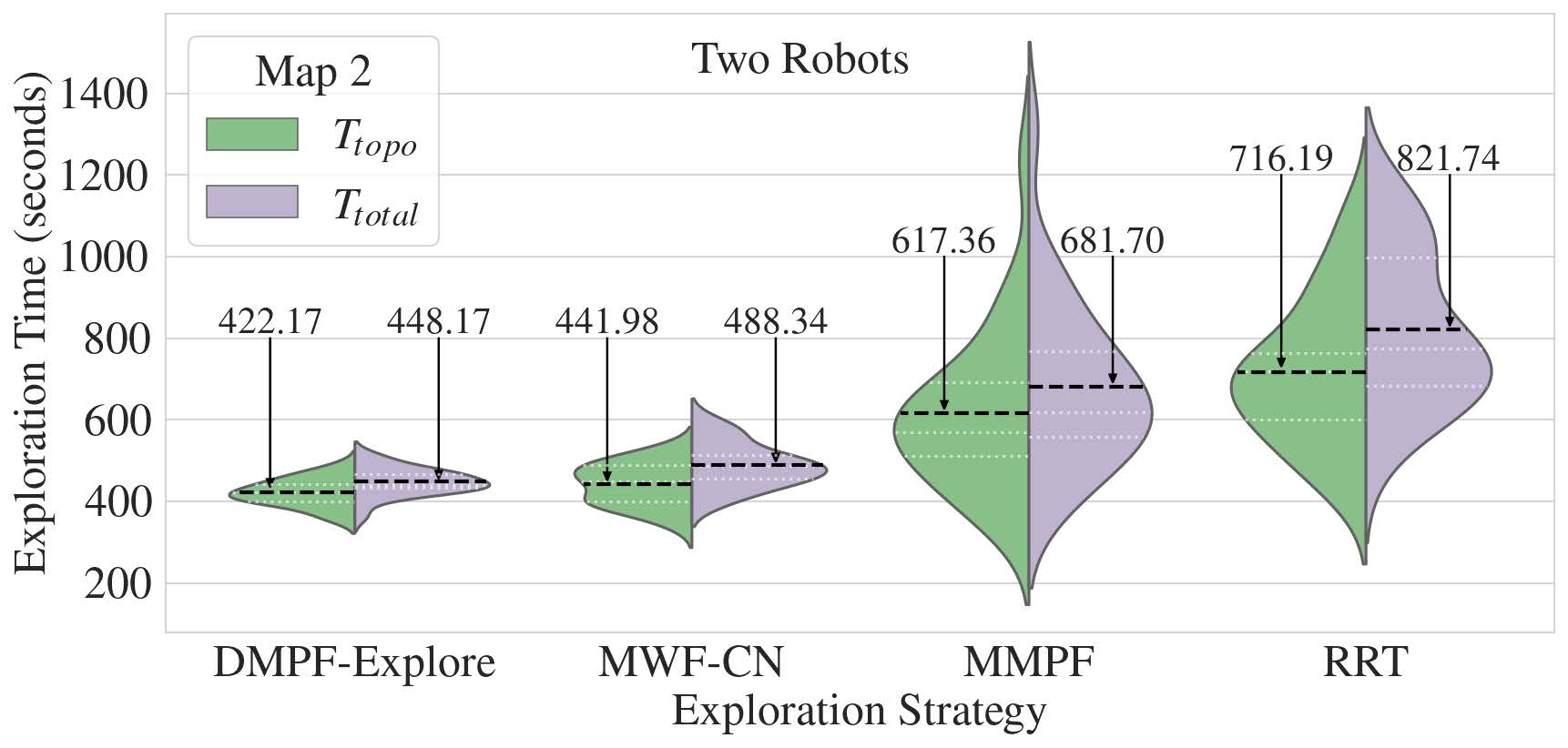}
         \caption{Two robots: $T_\text{topo}$ and  $T_\text{total}$ of Map 2}
         \label{fig:two_sim_map2_time}
     \end{subfigure}
     \begin{subfigure}[b]{0.4\linewidth}
         \centering
         \includegraphics[width=\linewidth]{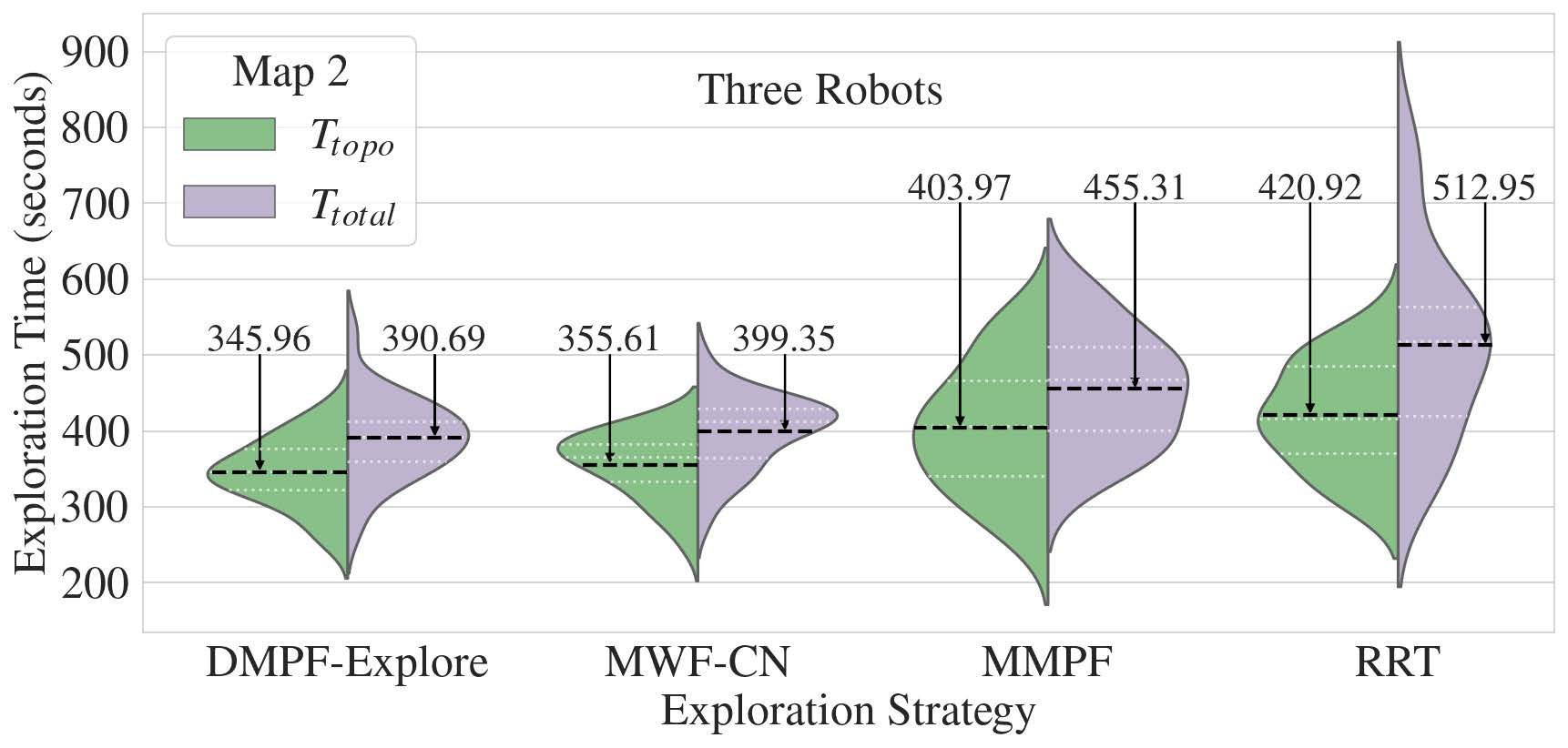}
         \caption{Three robots: $T_\text{topo}$ and $T_\text{total}$ of Map 2}
         \label{fig:three_sim_map2_time}
     \end{subfigure}
     \begin{subfigure}[b]{0.4\linewidth}
         \centering
         \includegraphics[width=\linewidth]{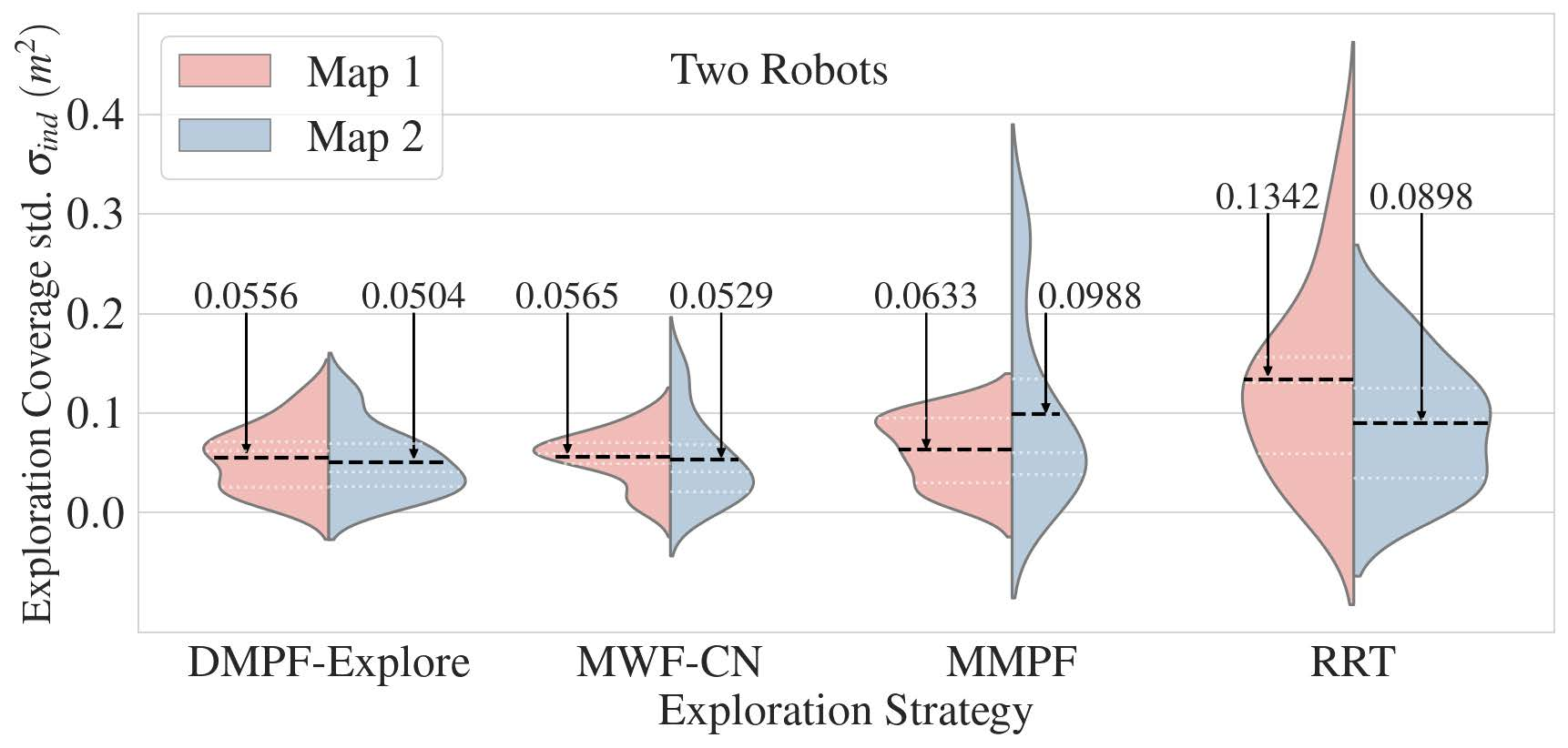}
         \caption{Two robots: $\sigma_\text{ind}$ of Map 1 and Map 2}
         \label{fig:two_sim_sigma}
     \end{subfigure}
     \begin{subfigure}[b]{0.4\linewidth}
         \centering
         \includegraphics[width=\linewidth]{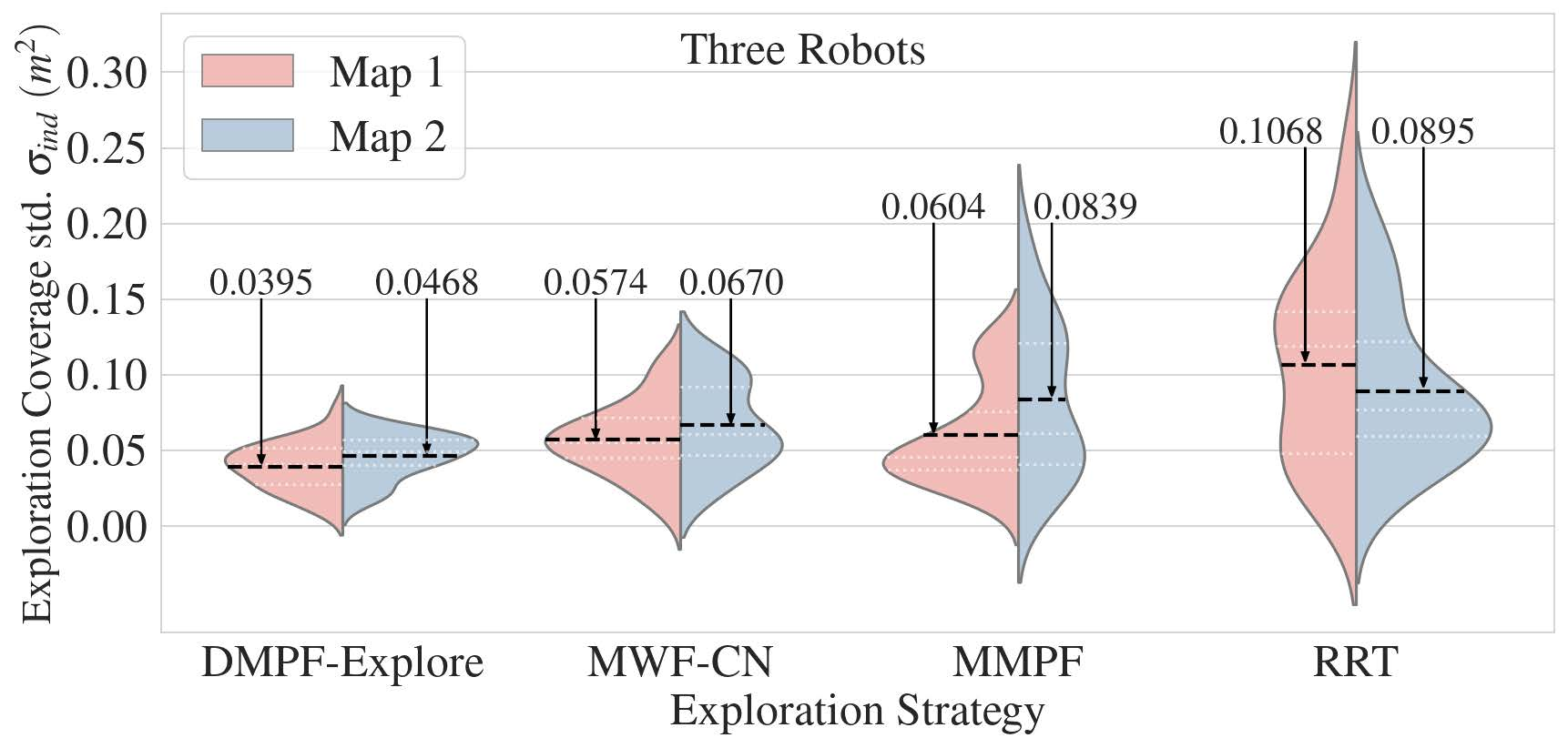}
         \caption{Three robots: $\sigma_\text{ind}$ of Map 1 and Map 2}
         \label{fig:three_sim_sigma}
     \end{subfigure}
     \begin{subfigure}[b]{0.4\linewidth}
         \centering
         \includegraphics[width=\linewidth]{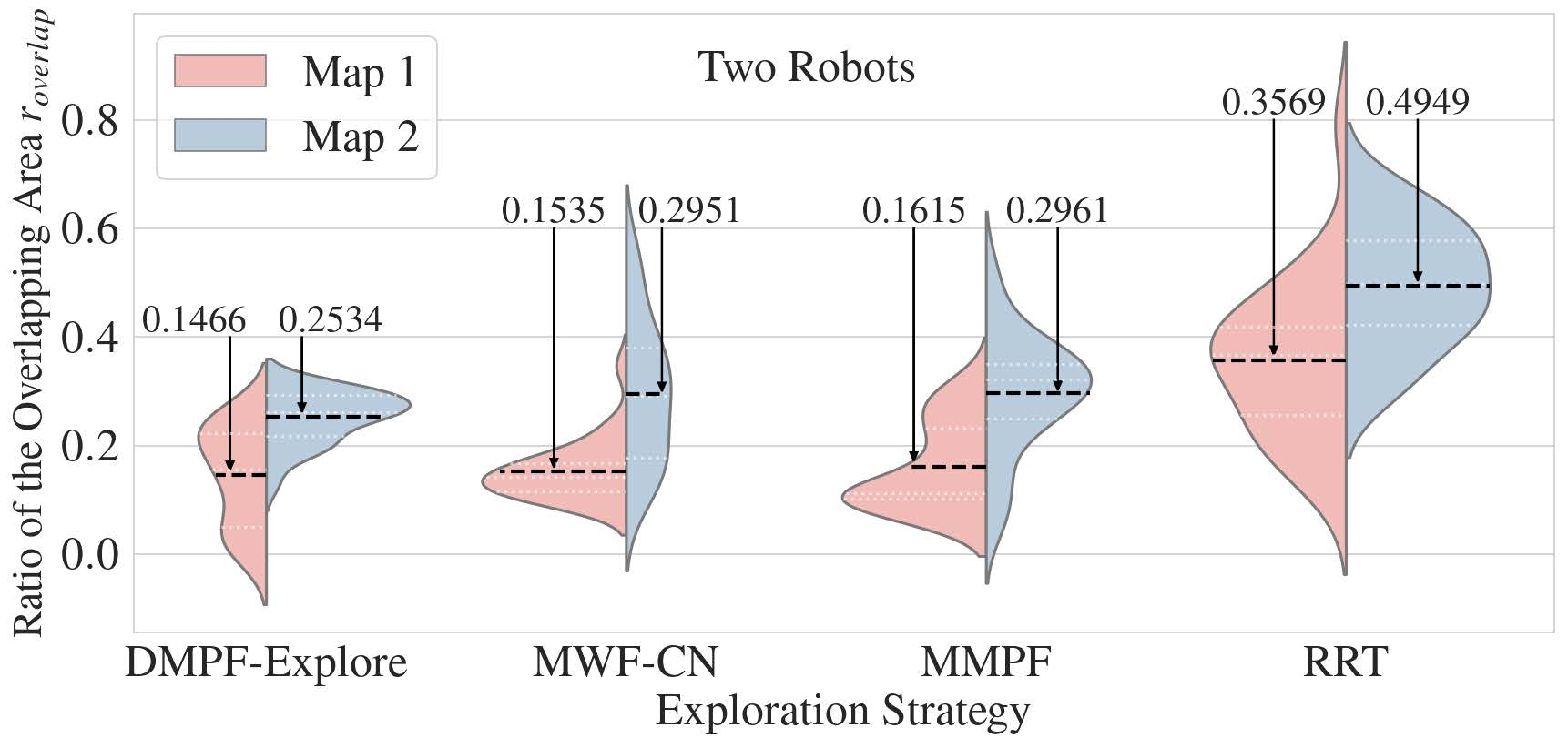}
         \caption{Two robots: $r_\text{overlap}$ of Map 1 and Map 2}
         \label{fig:two_sim_r_o}
     \end{subfigure}
     \begin{subfigure}[b]{0.4\linewidth}
         \centering
         \includegraphics[width=\linewidth]{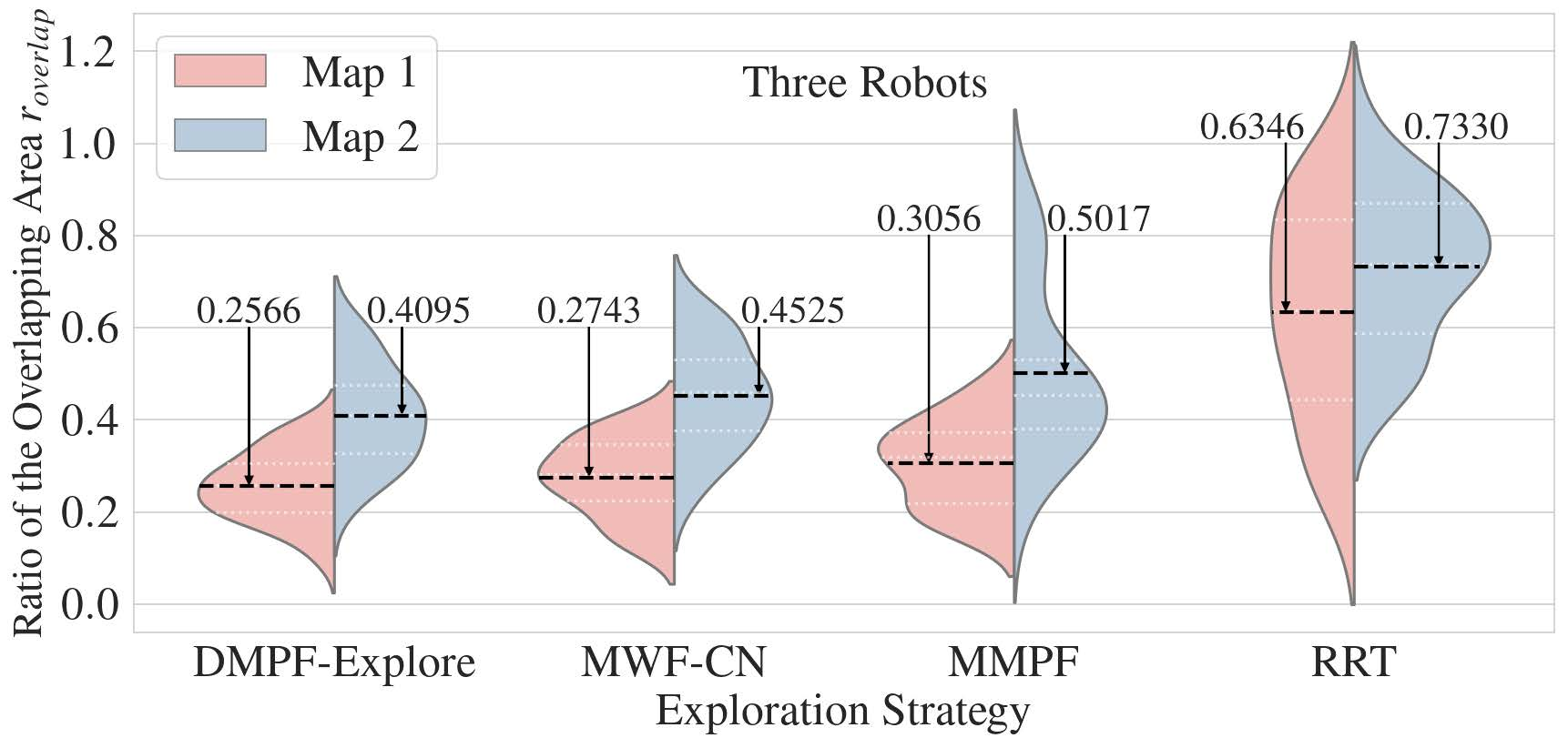}
         \caption{Three robots: $r_\text{overlap}$ of Map 1 and Map 2}
         \label{fig:three_sim_r_o}
     \end{subfigure}
        \caption{Violin plots for simulation results of DMPF-Explore, MWF-CN, MMPF, and RRT. The black dashed lines represent the mean values.}
        \label{fig:violin}
        \vspace*{-3mm}
\end{figure*}

\subsection{Parameter adjustment for the fastest exploration time}
Since we would like to have the best results of the MWF-CN, we investigate how appropriate parameters should be, and a more detailed account of this study is given in this subsection. Here we will explore which noise color $\alpha$ and variance $\sigma_d$ trigger the fastest exploration times $T_\text{topo}$ and $T_\text{total}$. The noise colors are white ($\alpha = 0$), pink ($\alpha = 1$), and brown ($\alpha = 2$). In addition, we fix $\sigma_r = 0.6$ in eq. (\ref{eq:P_r}) and select Map 2, as shown in Fig. \ref{fig:map2}, to be the environment for exploration. The benchmark's mapping \cite{Explore-Bench} is used as the mapping method, and we iterate 100 runs for each noise color.

According to the results in Fig. \ref{fig:optimal}, we can observe that the trend curves follow convex behavior. Each noise color $\alpha$ has different noise variance $\sigma_d$ intervals that produce the highest-speed exploration times $T_\text{topo}$ and $T_\text{total}$, as highlighted by the gray boxes. For the white noise ($\alpha = 0$), the variances $\sigma_d$ in $[0.035,0.065]$ give the fastest exploration times for both two and three robots, as shown in Fig. \ref{fig:two_alpha=0} and Fig. \ref{fig:three_alpha=0}. The variances $\sigma_d$ in $[0.020,0.050]$ perform better among other values of the pink noise ($\alpha = 1$) for both two and three robots, as shown in Fig. \ref{fig:two_alpha=1} and Fig. \ref{fig:three_alpha=1}. And finally, as shown in Fig. \ref{fig:two_alpha=2} for the case of two robots, the $\sigma_d = 0.095$ in the interval $[0.080,0.110]$ leads to the best performance. At the same time, as shown in Fig. \ref{fig:three_alpha=2} for the case of three robots, the $\sigma_d = 0.035$ in the interval $[0.020,0.050]$ triggers the best exploration times $T_\text{topo}$ and $T_\text{total}$ for the brown noise ($\alpha = 2$) and also better than the other noises.

This phenomenon is based on the nature of different noise behaviors. In this sense, the white noise is uncorrelated. On the other hand, because the brown noise is correlated, the values can be incremental at some periods. At the same time, the pink noise values are sometimes repeated, but there can also be some jumps. Since each noise variance, $\sigma_d$, also builds distinct intensities that can differently match the exploration conditions resulting in well-performing exploration times, the results in Fig. \ref{fig:optimal} can make sense for the aforementioned reasons. Although there can be a further investigation regarding this matter, these preliminary findings can logically guide which parameters should be used for the DMPF-Explore. Therefore, we select the brown noise ($\alpha = 2$) with $\sigma_d = 0.095$ for two robots and $\sigma_d = 0.035$ for three robots to apply with the MWF-CN and compare with the benchmark in the next subsection.

\subsection{Benchmark and comparison}
This subsection examines the effectiveness of the DMPF-Explore by comparing its performance to those of other exploration methods. Here the MWF-CN, the MMPF \cite{SMMR}, and the multi-robot version of RRT \cite{RRT, MultiRRT} will be utilized together with the benchmark's mapping. We conduct the simulation on two simulated environments shown in Fig. \ref{fig:maps}. Also, as mentioned previously, we will use the same set of parameters obtained from the preceding subsection, which are $\sigma_r=0.6$, $\alpha=2$, $\sigma_d=0.095$ for two robots, and $\sigma_d=0.035$ for three robots. For each method, the two-robot and three-robot explorations are run for 50 iterations, and we also evaluate the metrics introduced previously. The results are presented in Fig. \ref{fig:violin}. We can see that the DMPF-Explore performs the best compared to other methods in all simulated scenarios. First of all, comparing the DMPF-Explore to the MWF-CN with the benchmark's mapping, the results show that the DSMC-Map leads to better and more stable performance, which can be seen from the improvement in all metrics.
Also, as we can see from Fig. \ref{fig:two_sim_map1_time} to \ref{fig:three_sim_map2_time}, the robots using the DMPF-Explore can cover the overall structure of environments the fastest according to its lowest $T_\text{topo}$ (two robots: 105.63s for Map 1, 422.17s for Map 2, and three robots: 70.85s for Map 1, 345.96s for Map 2). It also shows the ability to complete the exploration areas at the highest speed, considering its $T_\text{total}$ (two robots: 120.92s for Map 1, 448.17s for Map 2, and three robots: 82.45s for Map 1, 390.69s for Map 2), which is the shortest time compared to other methods. In addition, it is apparent from the violin plots that the exploration times of the DMPF-Explore have the lowest dispersion, which can demonstrate the repeatability of our proposed method.
Considering the workload between robots, it can be seen from Fig. \ref{fig:two_sim_sigma} and \ref{fig:three_sim_sigma} that the robots using the DMPF-Explore have the most well-balanced exploration according to its variations and values of $\sigma_\text{ind}$ (two robots: 0.0556$\text{m}^2$ for Map 1, 0.0504$\text{m}^2$ for Map 2, and three robots: 0.0395$\text{m}^2$ for Map 1, 0.0468$\text{m}^2$ for Map 2). For the overlapping rate, we can observe from Fig. \ref{fig:two_sim_r_o} and Fig. \ref{fig:three_sim_r_o} that the results for three robots are higher than the ones for two robots. This increase occurs because the robots are more likely to repeat other robots' routes when we deploy a higher number of them. However, we find that the DMPF-Explore improves the ability of robots to avoid the mentioned situation, which can be inferred from its lowest $r_\text{overlap}$ (two robots: 14.66\% for Map 1, 25.34\% for Map 2, and three robots: 25.66\% for Map 1, 40.95\% for Map 2).

\begin{table}[ht]
\centering
\caption{Average computation time per cycle per robot of different exploration methods}
\label{tab:comput_time}
\begin{tabular}{|c|cccc|}
\hline
\multirow{3}{*}{\textbf{Environment}} & \multicolumn{4}{c|}{\textbf{\begin{tabular}[c]{@{}c@{}}Average computation time\\ per cycle per robot (ms)\end{tabular}}}                    \\ \cline{2-5} 
                                      & \multicolumn{2}{c|}{\textbf{MMPF}}                                              & \multicolumn{2}{c|}{\textbf{DMPF-Explore}}                 \\ \cline{2-5} 
                                      & \multicolumn{1}{c|}{\textbf{2 robots}} & \multicolumn{1}{c|}{\textbf{3 robots}} & \multicolumn{1}{c|}{\textbf{2 robots}} & \textbf{3 robots} \\ \hline
\textbf{Map 1}                        & \multicolumn{1}{c|}{705.81}            & \multicolumn{1}{c|}{763.62}            & \multicolumn{1}{c|}{711.74}            & 947.49            \\ \hline
\textbf{Map 2}                        & \multicolumn{1}{c|}{712.09}            & \multicolumn{1}{c|}{800.08}            & \multicolumn{1}{c|}{840.25}            & 998.36            \\ \hline
\end{tabular}
\end{table}

\subsection{Computation time analysis}
In this subsection, we delve into the computational aspects of the MMPF and our proposed method, DMPF-Explore, both of which demonstrate competitive exploration performance. Our first step is a comprehensive theoretical analysis of the time complexity, providing a brief understanding of the methods' time consumption. We then proceed to evaluate each robot's computation time per cycle during two-robot and three-robot exploration in two simulated environments. The results are presented in Table \ref{tab:comput_time}.

In terms of time complexity, since both methods are potential-field-based, the overall exploration time grows exponentially based on $N_C$ to the power of $S_\text{total}$, i.e., $\mathcal{O}\big(N_C^{S_\text{total}}\big)$, where $N_C$ is the number of frontier centroids and $S_\text{total}$ is the total exploration area. This complexity also aligns with the analysis in \cite{Potential-field-complexity}. For the attractive potentials of the MMPF and the DMPF-Explore, they are dominated by the original wave-front and MWF distances, respectively. Although their amounts of neighborhood sectors and distance weights are different, they do not affect the time complexities, which are $\mathcal{O}\big(S_\text{total}\big)$. The main element that the time complexities of the MMPF and our DMPF-Explore are noticeably different is repulsive potential. Since the MMPF utilizes the linear form, the complexity is only $\mathcal{O}(1)$. On the other hand, our DMPF-Explore, in which the potential is in exponential form and embedded with colored noise, requires $\mathcal{O}\big(S_\text{total}^2 \big)$ for the discrete Fourier transform \cite{FT-complexity} from noise generation.

As can be seen from Table \ref{tab:comput_time}, the computation time per cycle collected from each scenario also complies with our time complexity analysis, i.e., the MMPF, as expected, demands less time for each computation cycle than the DMPF-Explore. However, it is to be observed that less computation time does not inherently imply better exploration time since high-quality paths for robots can take longer time to compute. For example, according to Table \ref{tab:comput_time}, the computation time per cycle of the MMPF is 0.84\%-24.08\% less than the DMPF-Explore in Map 1 (two-robot: 705.81ms vs. 711.74ms, three-robot: 763.62ms vs. 947.49ms). Still, as shown in Fig. \ref{fig:two_sim_map1_time} and \ref{fig:three_sim_map1_time}, $T_\text{total}$ of the MMPF is 13.90\%-14.06\% slower than the DMPF-Explore (two-robot: 140.44s vs. 120.92s, three-robot: 95.94s vs. 82.45s). Similarly, for Map 2, the computation time per cycle of the MMPF is 18.00\%-24.78\% less than the DMPF-Explore (two-robot: 712.09ms vs. 840.25ms, three-robot: 800.08ms vs. 998.36ms). Yet, as presented in Fig. \ref{fig:two_sim_map2_time} and \ref{fig:three_sim_map2_time}, $T_\text{total}$ of the MMPF is also 14.19\%-34.26\% slower than the DMPF-Explore (two-robot: 681.70s vs. 448.17s, three-robot: 455.31s vs. 390.69s).

Furthermore, we can notice that the increases are not excessively high, and the overall computation times of the DMPF-Explore still do not surpass 1s. In prospective works, the relevant considerations can be investigated further to determine whether balancing the computation time and the exploration plan quality will lead to optimized performance.

\section{Real-world deployment}
\label{section:real-world}

\begin{figure}
    \centering
    \begin{subfigure}[b]{0.4\linewidth}
         \centering
         \includegraphics[width=\linewidth]{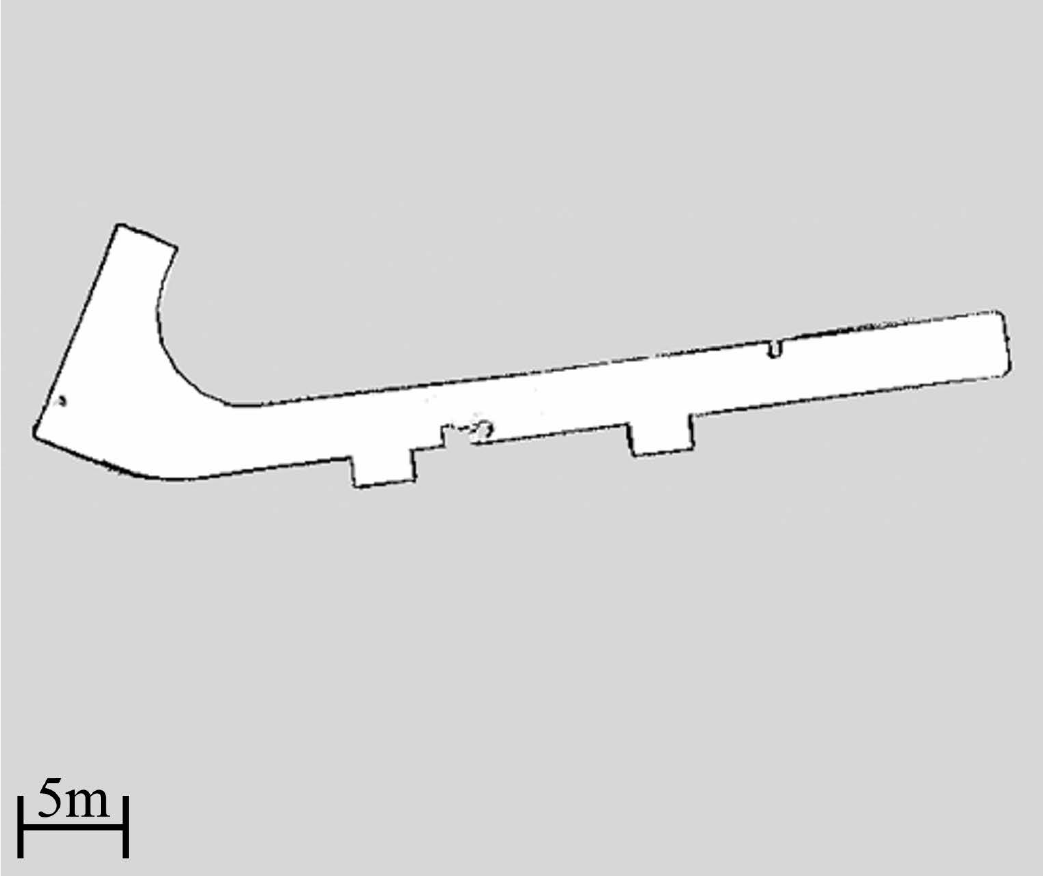}
         \caption{Real Environment 1}
         \label{fig:realmap1}
     \end{subfigure}
     \begin{subfigure}[b]{0.4\linewidth}
         \centering
         \includegraphics[width=\linewidth]{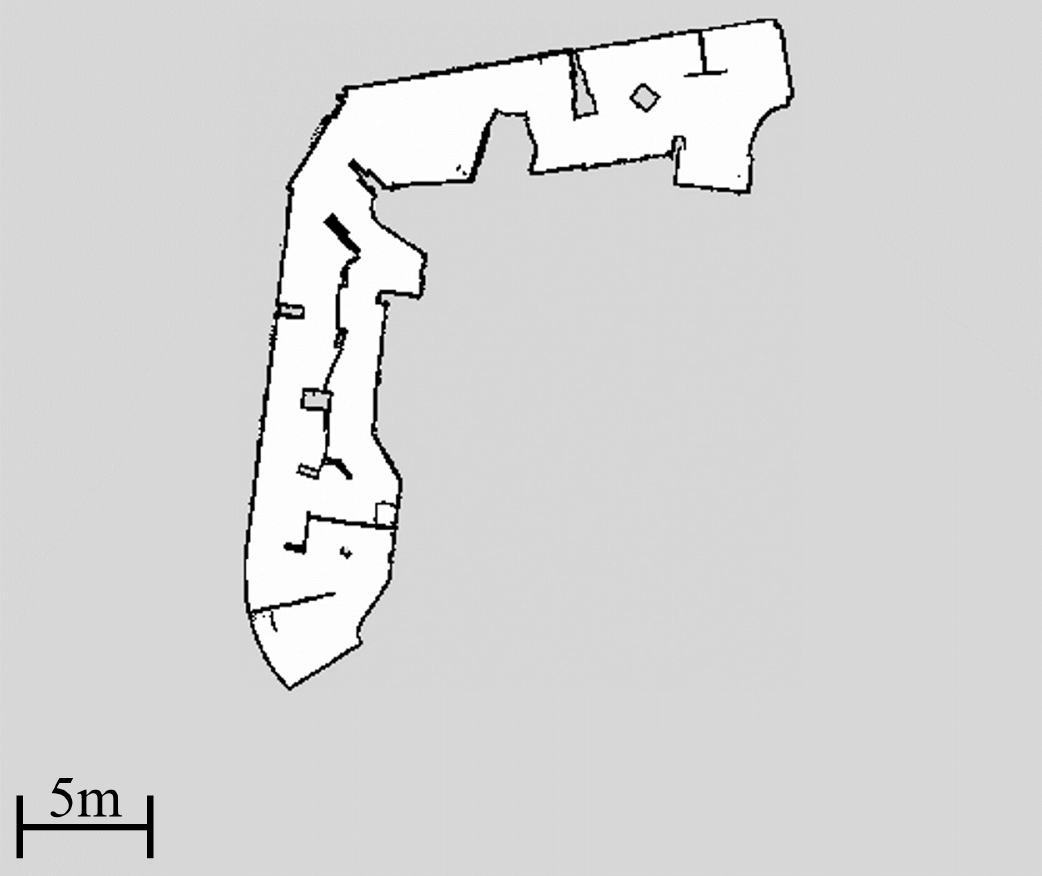}
         \caption{Real Environment 2}
         \label{fig:realmap2}
     \end{subfigure}
        \vspace*{-1mm}
        \caption{Real-world environments}
        \vspace*{-10mm}
        \label{fig:realmaps}
\end{figure}

\begin{figure}
     \centering
     \begin{subfigure}[b]{0.4\linewidth}
         \centering
         \includegraphics[width=\linewidth]{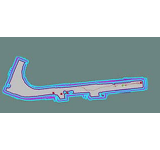}
         \vspace*{-11mm}
         \caption{MMPF used in Real Environment 1}
         \label{fig:rviz_realmap1_carto}
     \end{subfigure}
     \begin{subfigure}[b]{0.4\linewidth}
         \centering
         \includegraphics[width=\linewidth]{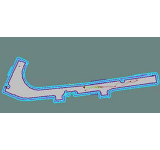}
         \vspace*{-11mm}
         \caption{DMPF-Explore used in Real Environment 1}
         \label{fig:rviz_realmap1_newmapping}
     \end{subfigure}
     \begin{subfigure}[b]{0.4\linewidth}
         \centering
         \includegraphics[width=\linewidth]{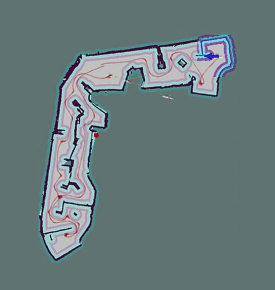}
         \caption{MMPF used in Real Environment 2}
         \label{fig:rviz_realmap2_carto}
     \end{subfigure}
     \begin{subfigure}[b]{0.4\linewidth}
         \centering
         \includegraphics[width=\linewidth]{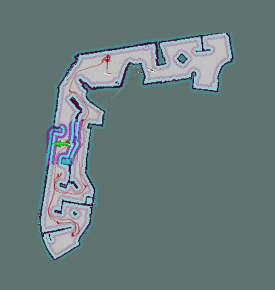}
         \caption{DMPF-Explore used in Real Environment 2}
         \label{fig:rviz_realmap2_newmapping}
     \end{subfigure}
        \caption{Maps generated by two-robot exploration using different methods. The red paths represent the robot's routes.}
        \label{fig:rviz_realmaps}
\end{figure}

According to the simulation results, we notice that our DMPF-Explore and the MMPF outperform the RRT. Hence, we employ and compare these methods for real-world deployment. The first subsection will provide details about which hardware is used and how the experiments are conducted. Then, the experimental results will be discussed.

\begin{figure*}
    \centering
    \begin{subfigure}[b]{0.8\linewidth}
         \centering
         \includegraphics[width=\linewidth]{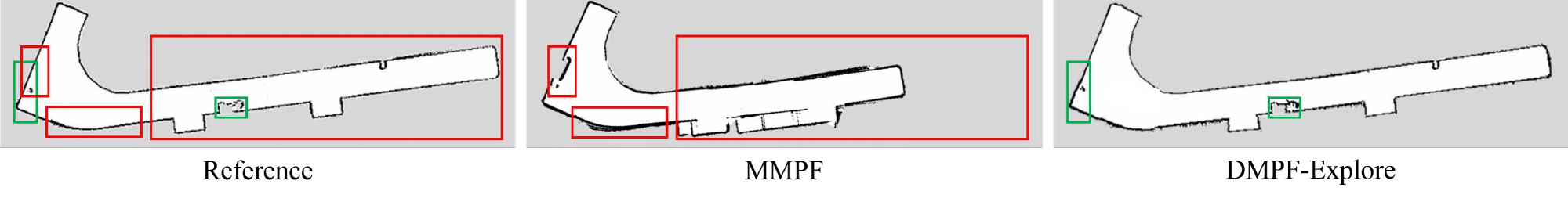}
         \caption{Real Environment 1}
         \label{fig:SSIM-realmap1}
     \end{subfigure}
     \begin{subfigure}[b]{0.8\linewidth}
         \centering
         \includegraphics[width=\linewidth]{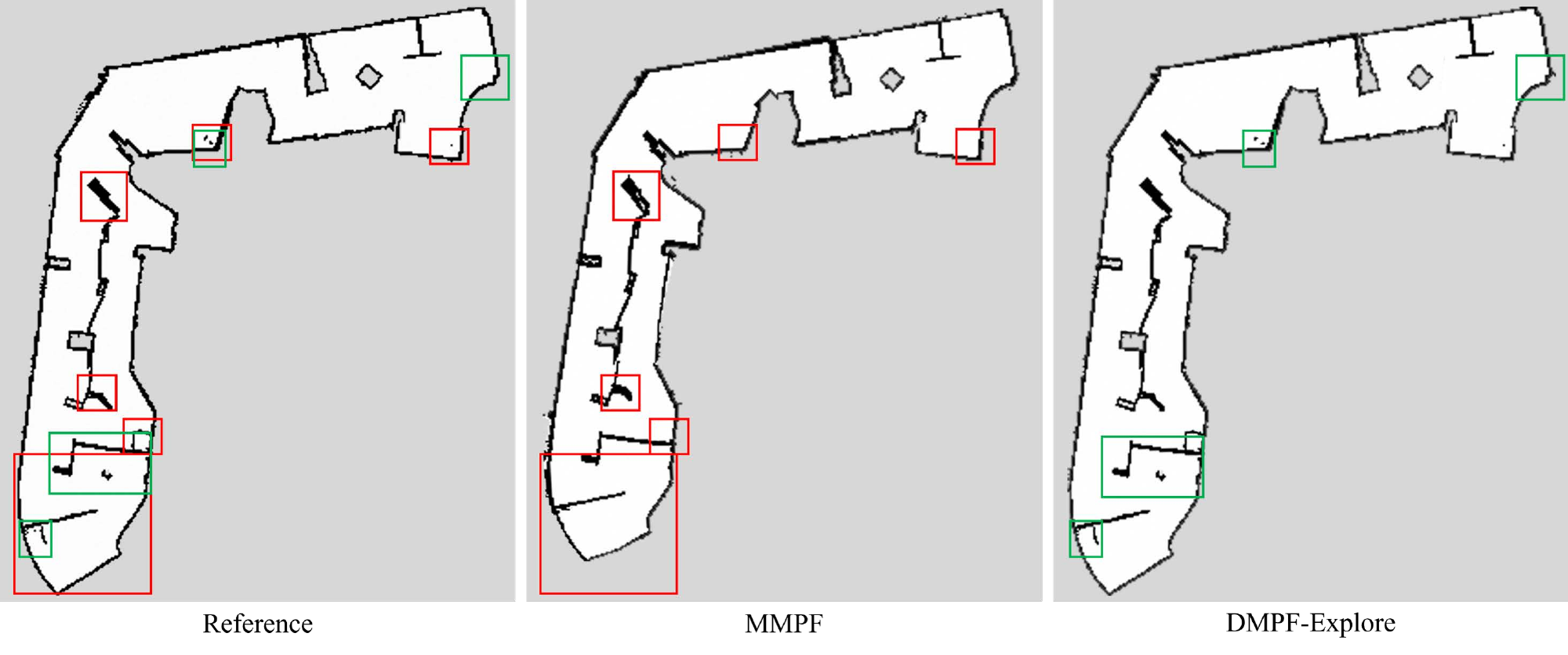}
         \caption{Real Environment 2}
         \label{fig:SSIM-realmap2}
     \end{subfigure}
        \caption{Example of the comparison between the reference and the resulting maps of each real environment by different methods. The red and green rectangles represent the differences between the reference map and the maps obtained from the MMPF and the DMPF-Explore, respectively.}
        \label{fig:SSIM}
\end{figure*}

\begin{table*}[ht]
\centering
\caption{Performance evaluation of two-robot exploration with different exploration methods in two environments}
\label{tab:real_results}
\begin{tabularx}{\textwidth}{ |c| *{6}{Y|} }
\cline{2-7}
   \multicolumn{1}{c|}{} 
 & \multicolumn{3}{c|}{\textbf{Real Environment 1}}  
 & \multicolumn{3}{c|}{\textbf{Real Environment 2}}\\
\hline
\textbf{Metrics} & \textbf{MMPF} & \textbf{DMPF-Explore} & \textbf{Improvement} & \textbf{MMPF} & \textbf{DMPF-Explore} & \textbf{Improvement}\\
\hline
$T_\text{topo}$ (s) & 209.48 & 121.11 & 72.97\% & 368.84 & 196.33 & 87.87\%\\
\hline
$T_\text{total}$ (s) & 217.06 & 156.35 & 38.83\% & 429.24 & 224.50 & 91.20\%\\
\hline
$\sigma_\text{ind}$ ($\text{m}^2$) & 0.1613 & 0.1289 & 0.0324 & 0.2783 & 0.0605 & 0.2178\\
\hline
$r_\text{overlap}$ (\%) & 64.26 & 40.42 & 23.84\% & 43.23 & 32.95 & 10.28\%\\
\hline
$R_\text{success}$ (\%) & 62.50 & 100.00 & 37.50\% & 83.33 & 100.00 & 16.67\%\\
\hline
$\text{SSIM}_\text{map}$ (\%) & 82.59 & 92.04 & 9.45\% & 86.57 & 95.55 & 8.98\%\\
\hline
\end{tabularx}
\vspace*{-1mm}
\end{table*}

\subsection{Hardware and experimental setup}

We deploy two TurtleBot2, run on a Kobuki base and equipped with a 2D LiDAR sensor with a maximum scanning range of 10m, as shown in Fig. \ref{fig:turtlebot}. Intel NUC i5 is used as the computing unit for each robot. And Intel NUC i7 is utilized for running visualization and evaluation scripts. We utilize the ASUS TUF-AX5400 router with WiFi 2.4GHz to cover the experiments' areas and connect the robots. In addition, the submap is published every 0.3 seconds. Regarding exploration environments shown in Fig. \ref{fig:realmaps}, the first is a long corridor of size 270$\text{m}^2$. The latter is an indoor area of size 218$\text{m}^2$ set up to be more complicated by using boxes and partitions. Both robots are spawned at the exact initial location in the middle of each environment. Note that to ensure exploration smoothness without any interruptions and to maintain the consistency of hardware setup without the requirements of the graphics processing unit (GPU) throughout all the experiments, the MMPF is exploited with the benchmark's mapping method, same as in \cite{Explore-Bench} instead of its mapping method from \cite{SMMR}.
For the DMPF-Explore, the same collection of parameters, which is [$\sigma_r=0.6$, $\alpha=2$, $\sigma_d=0.095$], is also applied to real robots. We also use the same evaluation metrics mentioned in Subsection \ref{subsection:metrics}.

\subsection{Experimental results and discussion}

Based on the results presented in Table \ref{tab:real_results}, for both environments, the DMPF-Explore performs significantly better than the MMPF. The robots using the DMPF-Explore can rapidly explore the overall exploration area resulting in 72.97\%–87.87\% enhancement in $T_\text{topo}$. The exploration task can also be completed fast as $T_\text{total}$ is 38.83\%–91.20\% more quickly. For the collaboration aspect, the robots using the DMPF-Explore have 0.0324–0.2178 lower in $\sigma_\text{ind}$, which means they have a more balanced workload allocation. At the same time, they also efficiently explore separate areas without too much overlapping, as its $r_\text{overlap}$ is 10.28\%–23.84\% lower. And finally, the DMPF-Explore leads to more successful exploration, which we can see 16.67\%–37.50\% raising in $R_\text{success}$. In addition, due to sensor noise and communication delay, more improvements are obtained in real-world deployment than in the simulation.

The multi-robot exploration performance using the DMPF-Explore is improved because of both map quality and exploration strategy. The resulting maps chosen from one robot are shown in Fig. \ref{fig:rviz_realmaps}. The DSMC-Map method provides maps that are accurate with the actual environments, while the benchmark's mapping method that is used for the MMPF exploration has challenges in accuracy and merging ability. For Real Environment 1, as shown in Fig. \ref{fig:SSIM-realmap1}, the map obtained from the MMPF can sometimes have wrong-merging parts because of its limitations in handling the empty environment with few features, while the one obtained from the DMPF-Explore is very precise. The average SSIM for the DMPF-Explore is 9.45\% better. Although the MMPF can perform better in Real Environment 2, as can be seen from Fig. \ref{fig:SSIM-realmap2}, many minor details can sometimes be neglected, unlike the map by the DMPF-Explore, which can accurately represent real environments. So, the average SSIM for the DMPF-Explore is 8.98\% higher. According to these results and Fig. \ref{fig:SSIM}, we can see that this development highlights the novelty of our DSMC-Map in improving map construction.

Regarding the exploration strategy aspect, the robots using the DMPF-Explore are not stuck because they have more extended neighborhoods for the attractive potential, and distinct diagonal and orthogonal weights also reduce the local optima occurrence. So, during the exploration, they are assigned to new accessible goals and can move further until they accomplish the whole exploration area. Also, we can see clearly from Fig. \ref{fig:rviz_realmap2_carto} that for the MMPF that has strict potentials, when one robot gets stuck, another robot has to be responsible for exploring the rest of the area. For some rounds, the robots using the MMPF also do not separate after starting exploration and continue to explore the same place. On the other hand, the robots using the DMPF-Explore travel in different directions and spend less time exploring because of their more flexible repulsive potential field and appropriate colored noise of the MWF-CN. The noise values embedded in the potential also help prevent the local optima issues. Therefore, together with the effective mapping method, DSMC-Map, the robots using the DMPF-Explore can explore the environments smoothly and successfully, resulting in better performance than the MMPF in all aspects. For clarification, a video of the experiments is also attached as supplementary material. 

\section{Conclusion and future work}
\label{section:conclusion}
This paper proposes the DMPF-Explore, a new distributed multi-robot exploration system consisting of a distributed mapping module, DSMC-Map, that provides accurate environment maps and the MWF-CN that can effectively deal with both simple and complicated map structures. In particular, the robots using the DSMC-Map can explore unknown environments efficiently. Furthermore, for the MWF-CN, the novel usage of the MWF distance and the colored noise augmented in the potential field increases the capability of robots to detect the goals during the exploration better and also strengthens the robots' adaptability in complex unknown environments. For the experiments, TurtleBots are deployed in two simulated environments of size 120$\text{m}^2$ and 391$\text{m}^2$ and two real environments of size 270$\text{m}^2$ and 218$\text{m}^2$. The results show that our system is more robust against the RRT and the MMPF for time efficiency and inter-robot collaboration in both simulation and real-world scenarios. There can be further studies to develop the exploration algorithm to be less sensitive to map incompleteness. Additional simulation and real-world experiments on unstructured open spaces would be beneficial to ensure the method's robustness in different environmental settings. Also, the computational efficiency can be considered as one of the evaluation metrics to show the method performance in another aspect. In addition, it is important to analyze the optimality of the colored noise used as a part of the repulsive potential further in order to guarantee its practicality in any unknown environment. Moreover, extending the DMPF-Explore for a higher number of robots and heterogeneous multi-robot systems is also an interesting topic for future work.

\section*{Declaration of competing interest}
The authors declare that they have no known competing financial interests or personal relationships that could have appeared to influence the work reported in this paper.

\section*{Data availability}
Data will be made available on request.

% Loading bibliography
\bibliographystyle{model1-num}
\bibliography{cas-refs}

\bio{Khattiya-photo-small}
\textbf{Khattiya Pongsirijinda} received the B.Sc. degree in Mathematics from Silpakorn University, Thailand, in 2019 and the M.Sc. degree in Data Science from the Skolkovo Institute of Science and Technology, Russia, in 2021. He was also a research student at the University of Nebraska–Lincoln, USA, in 2018. He is currently pursuing his Ph.D. degree at the Singapore University of Technology and Design, Singapore, under the supervision of Prof. Chau Yuen. His current research interests include robotics, multi-robot systems, and mathematical modeling. 
\endbio

\bio{Zhiqiang-photo-small}
\textbf{Zhiqiang Cao} received the B.S. and MA.Sc. degrees from the Southwest University of Science and Technology, Mianyang, China, in 2019 and 2022, respectively. He is currently pursuing his Ph.D. degree at the Singapore University of Technology and Design, Singapore, under the supervision of Prof. Chau Yuen. His current research interests include multi-robot systems, collaborative localization, and distributed SLAM systems.
\endbio

\vspace{2cm}

\bio{Kaushik-photo-small}
\textbf{Kaushik Bhowmik}
received the B.Tech. degree in Electronics and Instrumentation from Haldia Institute of Technology, India, in 2015, the M.Tech degree in Power Electronics and Electric Drives from IIT(ISM), India, in 2018, and the M.Sc. degree in Computer Control and Automation from Nanyang Technological University, Singapore, in 2019. He is currently pursuing his Ph.D. degree with the Inria Grenoble, Grenoble Alpes University, France. His current research interests include autonomous driving, computer vision, and robotics-related applications.
\endbio

\vspace{2cm}

\bio{Shalihan-photo-small}
\textbf{Muhammad Shalihan}
received the B.Eng. degree in Mechatronics from the University of Glasgow, Scotland, in 2020. He is currently pursuing his Ph.D. degree at the Singapore University of Technology and Design, Singapore, under the supervision of Prof. Chau Yuen. His current research interests include robotics, multi-robot and human-robot localization, and SLAM.
\endbio

\vspace{2cm}

\bio{Billy-photo-small}
\textbf{Billy Pik Lik Lau} received Ph.D. at the Singapore University of Technology and Design (SUTD) in 2021 as well as a degree in computer science and M. Phil degree in computer science from Curtin University in 2010 and 2014, respectively. He is currently a research fellow at SUTD Temasek lab. His Ph.D. research work includes Smart city, the internet of things, data fusion, urban science, and big data analysis. His current interest is the integration of data fusion, urban science, multi-agent system design, multi-robotic collaboration, and robotic exploration.
\endbio

\vspace{2cm}

\bio{Liuran-photo-small}
\textbf{Ran Liu} received the B.S. degree from the Southwest University of Science and Technology, China, in 2007, and the Ph.D. degree from the University of Tüebingen, Germany, in 2014. From 2015 to 2024, he was a research fellow at the Singapore University of Technology and Design. He is currently a senior research fellow at the Nanyang Technological University. His research interests include robotics and SLAM.
\endbio

\break

\bio{YC-photo-small}
\textbf{Chau Yuen} received the B.Eng. and Ph.D. degrees in information and communication from Nanyang Technological University, Singapore, in 2000 and 2004, respectively. He was a Post-Doctoral Fellow with Lucent Technologies Bell Labs, Murray Hill, in 2005, and a Visiting Assistant Professor with The Hong Kong Polytechnic University in 2008. From 2006 to 2010, he was with the Institute for Infocomm Research, Singapore. From 2010 to 2023, he was with the Singapore University of Technology and Design. Since 2023, he has been with the School of Electrical and Electronic Engineering, Nanyang Technological University. He is a Distinguished Lecturer of IEEE Vehicular Technology Society, Top 2\% Scientists by Stanford University, and also a Highly Cited Researcher by Clarivate Web of Science. He currently serves as an Editor-in-Chief for Springer Nature Computer Science, Editor for the IEEE Transactions on Vehicular Technology, IEEE System Journal, and IEEE Transactions on Network Science and Engineering. He also served as the Guest Editor for several special issues, including IEEE Journal on Selected Areas in Communications, IEEE Wireless Communications Magazine, IEEE Transactions on Cognitive Communications and Networking, and Elsevier Applied Energy.
\endbio

\newpage

\bio{UXuan-photo-small}
\textbf{U-Xuan Tan} received the B.Eng. and Ph.D. degrees from Nanyang Technological University, Singapore, in 2005 and 2010, respectively. From 2009 to 2011, he was a Postdoctoral Fellow with the University of Maryland, College Park, MD, USA. From 2012 to 2014, he was a Lecturer with the Singapore University of Technology and Design, Singapore. He took up a research-intensive role as an Assistant Professor in 2014. He has been promoted to an Associate Professor in 2021. His research interests include mechatronics, sensing and control, onsite micromanipulation and microsensing, sensing and control technologies for human–robot interaction, and interdisciplinary teaching.
\endbio

\end{document}